\documentclass[10pt,twocolumn,letterpaper]{article}

\usepackage{cvpr}              % To produce the CAMERA-READY version
\usepackage{xcolor}       % 提供颜色支持
\usepackage{listings}     % 核心宏包，用于排版代码
\usepackage[T1]{fontenc}  % 推荐的字体编码，以正确显示特殊符号

\usepackage{algorithm}
\usepackage{algorithmic}
\usepackage{booktabs, makecell, multirow, tabularx}
\usepackage{tcolorbox}
\tcbuselibrary{breakable,listings} % 加载tcolorbox的库

\usepackage{listings}
\usepackage{xcolor}
\usepackage{csquotes}

\definecolor{codegray}{rgb}{0.5,0.5,0.5}
\definecolor{codepurple}{rgb}{0.58,0,0.82}
\definecolor{codeblue}{rgb}{0,0,1}

\lstdefinestyle{mystyle}{
    commentstyle=\color{codegray},
    keywordstyle=\color{codeblue},
    stringstyle=\color{codepurple},
    basicstyle=\ttfamily\small, 
    breaklines=true,
    breakatwhitespace=true    
}
\newtcolorbox{promptbox}[1][]{
    breakable, 
    colback=black!5, 
    colframe=black!75, 
    fonttitle=\bfseries,
    title=#1, 
    arc=0mm, 
    boxrule=0.5pt, 
    left=3mm, right=3mm, top=2mm, bottom=2mm,
    listing only, 
    listing options={style=mystyle, language=text}, 
}

\definecolor{cvprblue}{rgb}{0.21,0.49,0.74}
\usepackage[pagebackref,breaklinks,colorlinks,allcolors=cvprblue]{hyperref}

\def\paperID{*****} % *** Enter the Paper ID here
\def\confName{CVPR}
\def\confYear{2026}

\title{Evo-Retriever: LLM-Guided Curriculum Evolution with Viewpoint-Pathway Collaboration for Multimodal Document Retrieval}

\author{
Weiqing Li,
Jinyue Guo,
Yaqi Wang,
Haiyang Xiao,
Yuewei Zhang$^{*}$,
Guohua Liu,
Hao Henry Wang$^{*}$\\
Alibaba Cloud Computing\\
{\tt\small \{liweiqing.lwq, liyou.zyw\}@alibaba-inc.com, cashenry@126.com}}

\begin{document}
\maketitle
\begingroup
\renewcommand\thefootnote{}\footnotetext{*Corresponding authors.}
\endgroup

\maketitle
\begin{abstract}

% ----------------- 20251111-----------------------
Visual-language models (VLMs) excel at data mappings, but real-world document heterogeneity and unstructuredness disrupt the consistency of cross-modal embeddings. Recent late-interaction methods enhance image-text alignment through multi-vector representations, yet traditional training with limited samples and static strategies cannot adapt to the model’s dynamic evolution, causing cross-modal retrieval confusion. To overcome this, we introduce \textbf{Evo-Retriever}, a retrieval framework featuring an LLM-guided curriculum evolution built upon a novel Viewpoint-Pathway collaboration. First, we employ multi-view image alignment to enhance fine-grained matching via multi-scale and multi-directional perspectives. Then, a bidirectional contrastive learning strategy generates "hard queries" and establishes complementary learning paths for visual and textual disambiguation to rebalance supervision. Finally, the model-state summary from the above collaboration is fed into an LLM meta-controller, which adaptively adjusts the training curriculum using expert knowledge to promote the model’s evolution. On ViDoRe V2 and MMEB (VisDoc), Evo-Retriever achieves state-of-the-art performance, with nDCG@5 scores of 65.2 \% and 77.1 \%. 

\end{abstract}    
\section{Introduction}
\label{sec:intro}

%-------- 2025 11 11 --------

Complex Visual Document Retrieval (CVDR) aims to precisely locate a relevant page or document from a large-scale corpus, based on a multimodal query that may reference text, layout, and visual elements~\cite{caffagni2025recurrence, xu2025multi}. Unlike traditional document retrieval, CVDR faces unique challenges due to the heterogeneous structure of documents such as financial reports, contracts, and scientific papers~\cite{lee2024unified, ma2024unifying}, where vital information is often dispersed across text, charts, and tables~\cite{gunther2025jina, meng2025vlm2vec}. Recent embedding-based methods tackle this by mapping the multimodal content of a document into a unified vector space~\cite{zhang2020deep, zhao2024controllable, wang2025vidorag, qin2025unimoco}. Among these, multi-vector late-interaction models such as ColPali~\cite{faysse2024colpali} achieve state-of-the-art (SOTA) performance, outperforming chunking-based ~\cite{karpukhin-etal-2020-dense,nassar2022tableformer,pfitzmann2022doclaynet} and single-vector approaches~\cite{radford2021learning, ma2024unifying}.

\begin{figure}
	\centering
	\includegraphics[width=80mm]{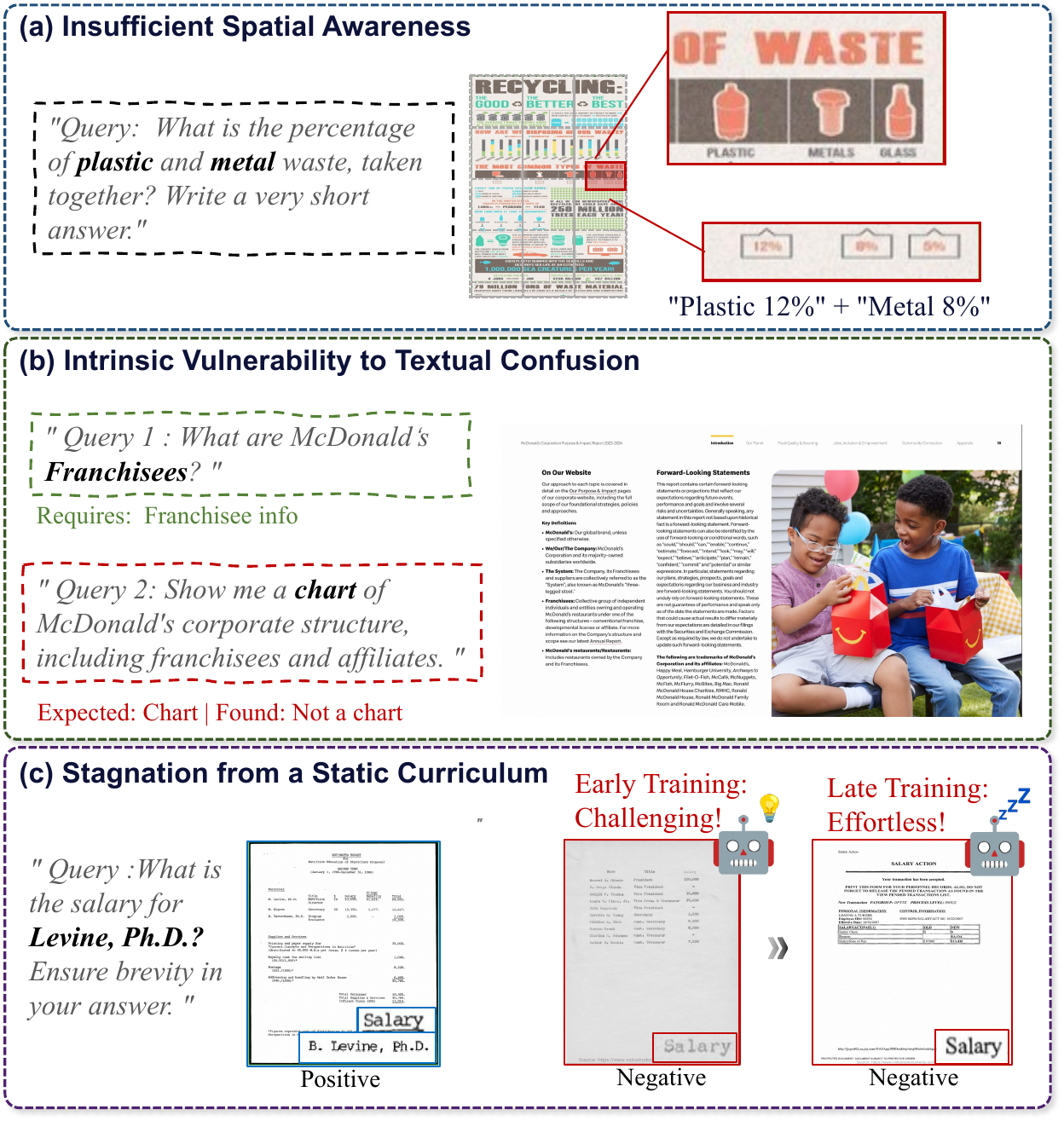}
	\caption{
        Challenges in complex visual document retrieval: 
        (a) \textbf{Insufficient Spatial Awareness}: Models must extract and integrate dispersed information (e.g., combining "Plastic 12\%" and "Metal 8\%") from dynamic layouts.
        (b) \textbf{Intrinsic Vulnerability to Textual Confusion}: Models must identify differentiated needs in obfuscated queries (e.g., franchisee info and corporate chart).
        (c) \textbf{Stagnation from a Static Curriculum}: The learning effectiveness of the initial negative sample set (e.g., containing only "salary") gradually diminishes under a fixed curriculum.
	}
	\label{fig_challenges}
\end{figure}

Despite recent progress, these advanced models suffer from three key limitations when dealing with complex real-world documents, as shown in Fig.~\ref{fig_challenges}. \textbf{First, they suffer from insufficient spatial awareness.} By relying on a single, fixed view, models struggle to integrate semantically related information points that are spatially distant. This results in a fragmented understanding of complex layouts~\cite{biswas2021graph, mahadevkar2024exploring, sourati2025lad, wang2025vrag}. \textbf{Second, they are intrinsically vulnerable to textual confusion.} Existing contrastive learning strategies primarily target hard negatives that are visually similar but semantically distinct, overlooking those that are textually similar yet visually mismatched~\cite{fu2023learning, huang2025visual, wang2025vidorag}.  \textbf{Third, and most importantly,  they are constrained by a static training curriculum.} Even SOTA methods that utilize techniques like data synthesis (GME~\cite{zhang2024gme}) typically rely on a fixed, pre-defined curriculum for selecting hard negatives. As the model's capabilities improve, it quickly masters the initial set of hard negatives. However, the static strategy continues to provide these now-unchallenging samples, which ultimately hinders the model's discriminative ability~\cite{teiletche2025modernvbert, he2023capstone, kwak2025qure, jang2023difficulty}.

%第三，它们受限于一种静态的训练课程。 即使是当前最先进的方法，无论是利用数据合成技术（如GME [18]）还是高级采样策略（如Nomic Embed [16]），也通常依赖一个固定的、预定义的课程来选择难负样本。随着模型能力的提升，它会迅速掌握最初设定的难负样本，但静态策略依旧持续提供这些已失去挑战性的样本，这最终会阻碍模型判别能力的进一步提升。(GME\cite{zhang2024gme}) or advanced sampling (Nomic Embed\cite{nomicembedmultimodal2025})

\begin{figure}
	\centering
	\includegraphics[width=80mm]{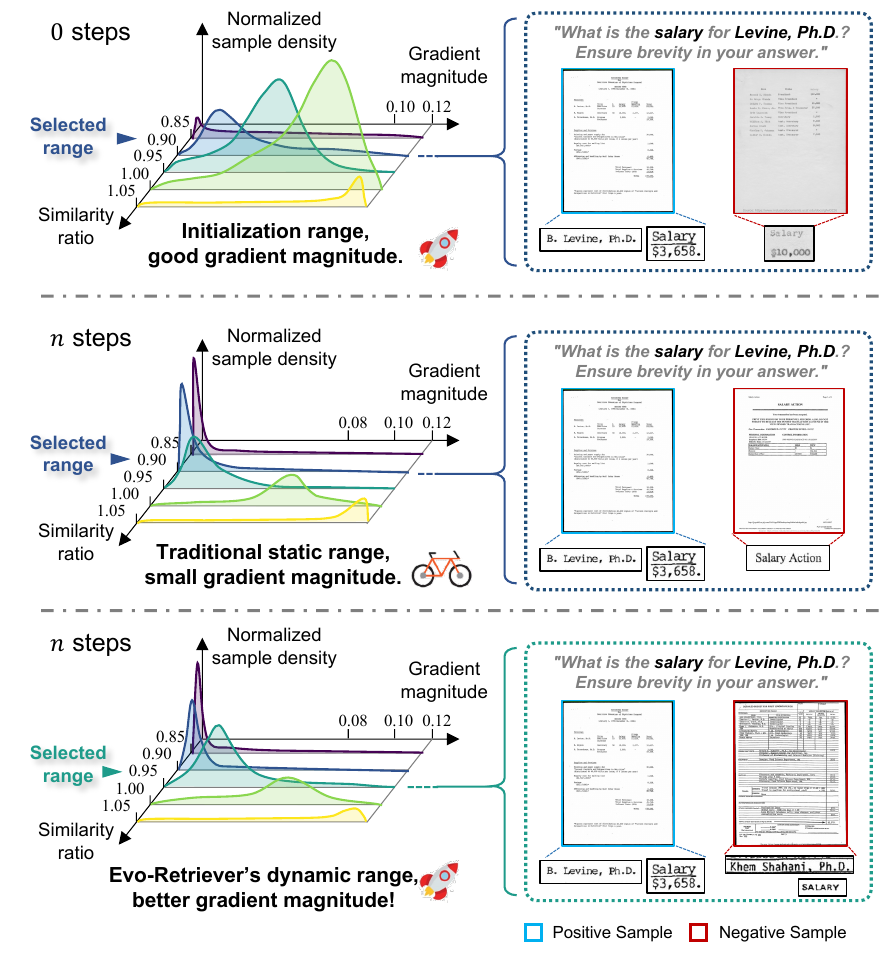}
	\caption{
            The motivation behind LLM-guided 
evolutionary curriculum. 
            (a) A static threshold is effective in the early training stage, providing informative gradient signals. 
            (b) As the model converges, the same threshold captures only trivial negative samples, leading to diminished gradients. 
            (c) Our LLM-guided evolutionary curriculum adjusts the mining interval to continually discover challenging negatives, ensuring sustained optimization.
            }
	\label{fig_motivation}
\end{figure}

Therefore, we propose \textbf{Evo-Retriever}, a retrieval framework where the model and the training curriculum co-evolve to address the above challenges. It integrates three components. First, to enhance spatial perception, we introduce \textbf{Multi-view Image Alignment (MVA)} to construct multi-perspective views of each document via rotation, scaling, and stitching. These views, together with the original, are processed in parallel by a shared visual encoder and are jointly aligned with the text embedding through a consistency loss, thereby enforcing representation consistency across multiple views. Second, \textbf{Bidirectional Contrastive Learning (BCL)} augments conventional query-to-document hard-negative mining with a reverse document-to-query path. A reverse sample generator synthesizes textually similar but visually distinct hard negative queries, enabling the model to anchor semantics to relevant visual evidence and improving robustness against textual perturbations. Finally, under the \textbf{Viewpoint–Pathway Collaboration} formed by MVA and BCL, the \textbf{LLM-guided Evolutionary Curriculum (LLM-EC)} replaces the static curriculum arrangement with dynamic stage regulation. An external LLM meta-controller analyzes training-state summaries (e.g., loss dynamics and historical decision performance) to adaptively adjust the difficulty of hard negatives (see \cref{fig_motivation}), ensuring efficient supervision.

%1105 LLM-guided dynamic curriculum
%最后，也是最关键的一点，我们提出由外部 LLM 引导的进化课程。embedding模型的训练过程被划分为多个阶段，LLM作为一个元控制器，在每一个阶段内分析训练状态的摘要——例如损失趋势和历史决策表现——以自适应地调度难负样本的难度，确保训练方案在检索模型进化的同时保持挑战性和高效性。

%Third, we design a staged dynamic hard negative mining mechanism. At each stage, the trained model serves as a teacher to re-sample challenging negatives for each query, while adaptive thresholds ensure that sample difficulty evolves alongside model capacity.

\vspace{0.5em}
\noindent\textbf{Our contributions can be summarized as follows:}
\begin{itemize}

    \item We introduce Evo-Retriever, a retrieval framework for complex visual documents based on a \textbf{model-curriculum co-evolution} paradigm.

    \item The \textbf{MVA-BCL} combination improves representation robustness by enhancing spatial invariance and reducing textual confusion.

    \item The \textbf{LLM-EC} adaptively schedules hard-negative difficulty based on training-state feedback, ensuring supervision remains challenging during model evolution.

    \item Evo-Retriever\footnote{Model checkpoints:
https://huggingface.co/ApsaraStackMaaS} sets new SOTA across several benchmarks, achieving nDCG@5 of 65.20 \% on ViDoRe V2 and 77.12 \% on MMEB.
%(+1.70 \% over SOTA)
% (+0.32 \% over SOTA)
\end{itemize}

\section{Related works}

\paragraph{Paradigm Shift in Document Retrieval: From Text Parsing to Visual Encoding.}
Traditional retrieval pipelines~\cite{karpukhin-etal-2020-dense,nassar2022tableformer} rely on a ``parse-then-retrieve'' approach. This requires complex parsing, OCR, and chunking steps that are error-prone and disrupt the page's original layout and visual context, leading to the loss of crucial cues from tables, charts, and fonts~\cite{pfitzmann2022doclaynet}. To circumvent these issues, methods like DSE~\cite{ma2024unifying} treat page screenshots as a unified input, directly encoding them into dense vectors using a VLM-based bi-encoder. However, a single global vector limits their fine-grained alignment and local discriminative capabilities~\cite{monsefi2024detailclip}.

\paragraph{Refining Dense Representations: From Global Matching to Multi-Grained Interaction.}
To address the granularity limitations of a single vector, ColBERT~\cite{khattab2020colbert} proposed the late interaction paradigm, which indexes documents at the token level and computes pairwise similarities with query tokens at runtime for fine-grained alignment. This idea has been successfully transferred to the document image domain. ColPali~\cite{faysse2024colpali} generates multi-vector representations for pages based on a VLM, significantly outperforming text-centric and single-vector approaches on page-level retrieval benchmarks like ViDoRe~\cite{macé2025vidorebenchmarkv2raising}. The efficacy of this approach has recently been confirmed by numerous state-of-the-art models, including Llama-Nemoretriever~\cite{xu2025llama}, as well as strong open multimodal embedding models such as colnomic-embed-multimodal-7b. Meanwhile, the fusion of PaLI-style architectures~\cite{chen2022pali} and OCR-free VLMs~\cite{liu2024textmonkey, nacson2025docvlm} further underscores the necessity of reading textual and visual semantics directly from pixels while preserving layout signals. However, long-range dependencies and cross-region information aggregation remain challenges~\cite{caffagni2025recurrence}. Our work follows the late interaction paradigm with an OCR-free visual backbone and enhances alignment stability under layout variations and spatial fragmentation by introducing a multi-view consistency constraint.

\paragraph{Optimizing Training Strategies for Retrieval.} High-quality negative samples are crucial for retrieval, leading to a variety of sophisticated training strategies. Data-centric efforts range from dataset curation (CoRNStack~\cite{suresh2024cornstack}) and synthesis (GME~\cite{zhang2024gme}) to advanced hard negative mining, including in-batch methods and LLM-judge-based approaches (UniME-V2~\cite{gu2025unime}). Others refine the training process itself, through novel objective functions (CAFe~\cite{yu2025cafe}, QQMM-Embed~\cite{xue2025improve}) or elaborate multi-stage pipelines (Llama-Nemoretriever~\cite{xu2025llama}, Seed-1.6-Embedding, UME-R1~\cite{lan2025ume}). Different from prior work, our method introduces a new learning path and a dynamic curriculum. First, we establish a document-{to}-query path and leverage a VLM to synthesize "textually similar, visually mismatched" negatives to enhance the model's semantic disambiguation. Second, we design a dynamic hard-negative mining curriculum where, through adaptive thresholds and model refreshing, the difficulty of negatives co-evolves with the model's capabilities.
\section{Methodology}

\begin{figure*}
	\centering
	\includegraphics[width=166mm]{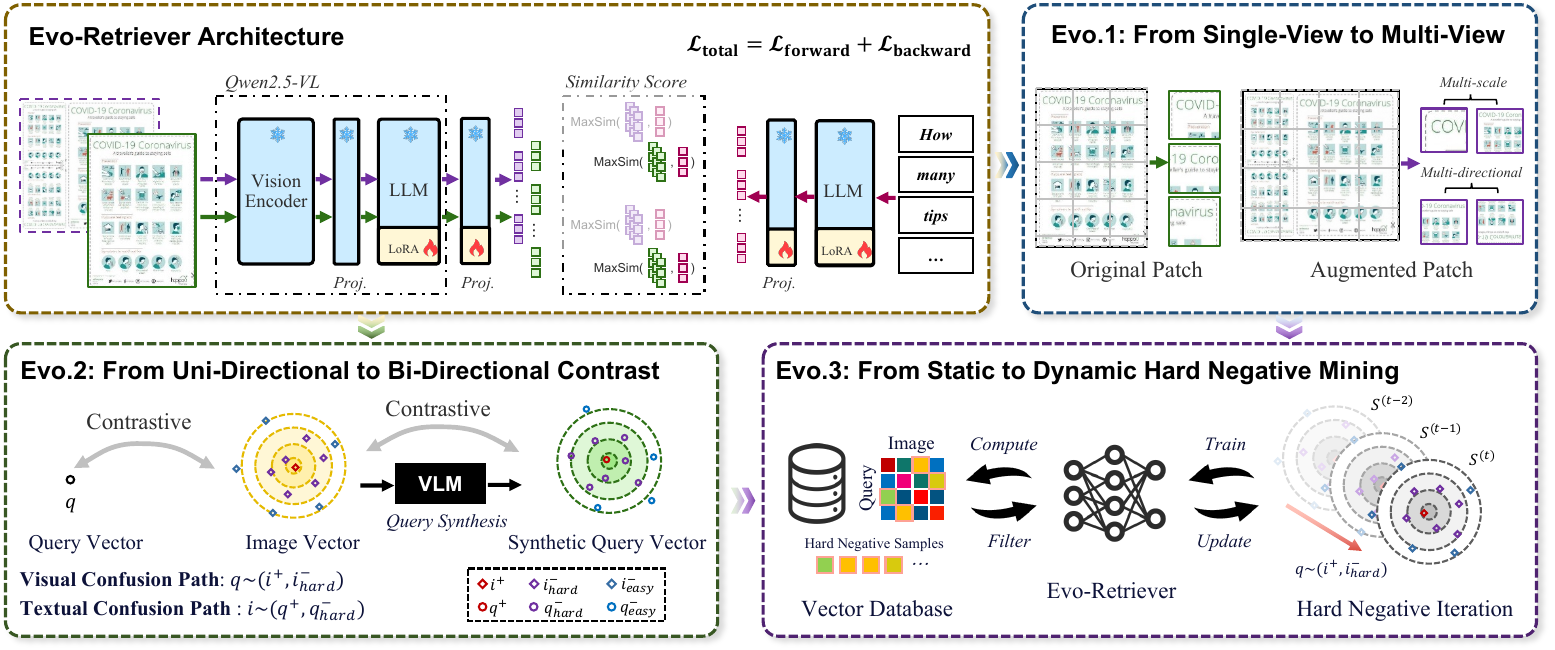}
	\caption{
		% Overview of the proposed EvoQwen-VL: \textcolor{red}{...}
        Evo-Retriever enhances representation via multi-view (Evo. 1) and bidirectional contrast (Evo. 2). Guided by expert knowledge, this Viewpoint-Pathway collaboration dynamically mines hard samples (Evo. 3), enabling continual evolution of the model.
	}
	\label{fig_overview}
\end{figure*}

As shown in \cref{fig_overview}, Evo-Retriever builds on a visual–language backbone ~\cite{bai2025qwen2} with multi-vector late interaction ~\cite{khattab2020colbert}. It fuses three synergistic components: (i) MVA and (ii) BCL, forming the Viewpoint-Pathway collaboration that strengthens spatial perception and mitigates textual ambiguities, and (iii) an LLM-EC meta-controller that adaptively tunes hard-negative mining based on training-state summaries to promote continuous model evolution.

\subsection{Representation Enhancement via Viewpoint-Pathway Collaboration}
This module combines two synergistic components: MVA to strengthen spatial perception (Viewpoint) and BCL to mitigate textual confusion (Pathway). Together, they bolster the model’s foundational representations, enabling continual learning under an evolving curriculum and supporting the retriever’s sustained representational evolution.

% \paragraph{{Multi-View Alignment} (MVA).} \label{sec:MVA} To address insufficient spatial awareness, we introduce the MVA strategy, which enforces robustness to minor geometric transformations (scale and rotation) and stabilizes query similarity scores for semantically identical documents.

% As shown in Fig. \ref{fig_overview}, for each image $I$, we construct a \emph{multi-view composite} $I^{\text{aug}}$ by horizontally concatenating a down-sampled version and a rotated version within $[-180^\circ, 180^\circ]$. Coupled with smart resizing of the Qwen-VL backbone, the concatenated image $I^{\text{aug}}$ breaks the fixed patching regime within the original token budget. It shares a common mapping path with $I$, aligning with the query $Q$ it matched.

% Training employs two alignment losses, $\mathcal{L}_{Q \to I}$ and $\mathcal{L}_{Q \to I^{\text{aug}}}$, integrated into the overall objective to encourage invariance across scales and orientations under the same textual query. The MVA strategy is applied only during training and does not add inference cost, enhancing robustness for multimodal document retrieval and natural image retrieval.

\paragraph{{Multi-View Alignment} (MVA).} \label{sec:MVA} To address insufficient spatial awareness, we introduce the MVA strategy, which enforces robustness to minor geometric transformations (scale and rotation) and stabilizes query similarity scores for semantically identical documents. Using only a single augmentation often distorts document layout and may corrupt fine-grained visual cues, causing the model to overfit to the perturbed view, whereas our hybrid design preserves the original view while introducing controlled geometric variation.

% As shown in Fig. \ref{fig_overview}, for each image $I$, we construct a \emph{multi-view composite} $I^{\text{aug}}$ by horizontally concatenating a down-sampled version and a rotated version within $[-180^\circ, 180^\circ]$. Coupled with smart resizing of the Qwen-VL backbone, the concatenated image $I^{\text{aug}}$ breaks the fixed patching regime within the original token budget. It shares a common mapping path with $I$, aligning with the query $Q$ it matched. 

As shown in Fig.~\ref{fig_overview}, for each image $I$, we build a \emph{multi-view composite} $I^{\text{aug}}$ by horizontally concatenating the original image, a down-sampled version, and a rotated version with angles sampled from $[-180^\circ, 180^\circ]$. Leveraging Qwen-VL's smart resizing, the composite yields a different patch layout within the same token budget. Both $I$ and $I^{\text{aug}}$ share the same mapping pathway and align with the matched query $Q$.

% As shown in Fig.~\ref{fig_overview}, for each image $I$, we construct a \emph{multi-view composite} $I^{\text{aug}}$ by horizontally concatenating the original image, a down-sampled version, and a rotated version with the rotation angle sampled from $[-180^\circ, 180^\circ]$. Leveraging the smart resizing mechanism of the Qwen-VL backbone, the resulting composite image produces a different patch layout while remaining within the original token budget. Both $I$ and $I^{\text{aug}}$ share the same mapping pathway and are aligned with the matched query $Q$.

Training employs two alignment losses, $\mathcal{L}_{Q \to I}$ and $\mathcal{L}_{Q \to I^{\text{aug}}}$, integrated into the overall objective to encourage invariance across scales and orientations under the same textual query. The MVA strategy is applied only during training and does not add inference cost.

\paragraph{{Bidirectional Contrastive Learning} (BCL).}
\label{sec:BCL} Mainstream multimodal retrievers rely on a uni-directional query-to-document (Q$\to$D) contrastive objective, which neglects the document-to-query (D$\to$Q) path and leaves textual confusion underexplored. We adopt a bi-directional learning framework with contrastive constraints on both directions. Unlike conventional methods that construct hard negative queries for the D$\to$Q path using simple in-batch negatives, we draw inspiration from DocReRank~\cite{wasserman2025docrerank} to introduce an automated \textbf{Hard Negative Query Synthesis (HNQS)} Pipeline. It generates diverse negatives that are syntactically or contextually similar to the positive query but semantically contradict the associated image.

Specifically, for a positive image-query pair $(I_{pos}, Q_{pos})$, the HNQS pipeline generates a set of negative queries {$\{Q_{neg}\}$} that are syntactically or contextually similar to $Q_{pos}$ but semantically contradictory to the content of image $I_{pos}$. The HNQS pipeline generates a candidate pool of negative queries for each positive pair, which contains 20 candidates in our experiments. From this pool, our LLM-EC strategy (detailed in Sec.~\ref{sec:llm_curriculum}) then dynamically selects negatives during training to form the contrastive sample.

The prompt design follows DocReRank, leveraging strong vision–language understanding and generation capabilities (details in the Appendix).
%(details in Appendix \cref{sec:NegativeQueryGenerationPrompts}
%.

These carefully crafted negative queries are essential to the success of the subsequent bi-directional contrastive learning, as they compel the model to go beyond surface-level text matching and achieve deep semantic alignment with visual content. 

\paragraph{{Overall Optimization Objective}.} \label{sec:OverallObjective} We unify the above two enhancement components into a single optimization objective using a \emph{softplus-based margin loss}, which fully leverages mined and synthesized hard negatives. 
Unlike softmax-normalized InfoNCE~\cite{oord2018representation}, this margin loss treats each hard negative independently, ensuring consistent gradient signals and progressively sharpening decision boundaries. The overall loss is defined as:
\begin{equation}
    \mathcal{L}_{total} = \mathcal{L}_{\text{forward}} + \alpha \cdot \mathcal{L}_{\text{backward}},
    \label{eq:overall_loss}
\end{equation}
where $\alpha$ is a hyper-parameter controlling the relative contributions of the two loss components.

The \textbf{forward contrastive loss}, $\mathcal{L}_{\text{forward}}$, incorporating MVA, is a dual-view objective. It comprises two contrastive terms: one computed on the original view (using $I_{\text{ori}}$ and its negatives $\{I_{\text{neg}}\}$) and another on the augmented view (using the augmented positive image $I^{\text{aug}}_{\text{ori}}$ and the corresponding augmented negatives $\{I^{\text{aug}}_{\text{neg}}\}$). Both terms are contrasted against the same query $Q_{\text{pos}}$:
%%
%$I^{\text{aug}}_{\text{neg}}$
%$I^{\text{aug}}_{\text{ori}}$
%
\begin{equation}
\label{eq:forward_loss_revised}
\mathcal{L}_{\text{forward}} = \mathcal{L}(Q_{\text{pos}}, I_{\text{ori}}, \{I_{\text{neg}}\}) + \beta \cdot \mathcal{L}(Q_{\text{pos}}, I^{\text{aug}}_{\text{ori}}, \{I^{\text{aug}}_{\text{neg}}\})
\end{equation}
where $\beta$ is a balancing hyperparameter, and $\{I_{neg}\}$ is a set of hard negative documents dynamically sampled as detailed in Sec.~\ref{sec:llm_curriculum}. The generic margin loss function $\mathcal{L}(\cdot)$ is defined as:
\begin{equation}
    \begin{split}
        \mathcal{L}(Q, I_{pos}, \{I_{neg}\}) = \sum_{k=1}^{K} \log(1 + \exp( \\
        \frac{\text{sim}(Q, I_{neg}^{(k)}) - \text{sim}(Q, I_{pos})}{\tau})),
    \end{split}
    \label{eq:margin_loss_definition}
\end{equation}
where  $\tau$ is a temperature hyperparameter. 
The late-interaction score $\text{sim}(Q, I)$ between a query $Q$ and an image $I$ is calculated based on their respective token embeddings. Let $E_Q(Q) = \{q_1, \dots, q_{L_Q}\}$ and $E_I(I) = \{i_1, \dots, i_{L_I}\}$ be the sets of token embeddings, where $L_Q$ and $L_I$ represent the number of tokens in the query and image, respectively, and each token $q_l, i_j \in \mathbb{R}^d$. Following ColBERT~\cite{khattab2020colbert}, the score is:
\begin{equation}
    \text{sim}(Q, I) = \sum_{l=1}^{L_Q} \max_{j=1}^{L_I} \left( E_Q(Q)_l \cdot E_I(I)_j^T \right),
    \label{eq:late_interaction_score}
\end{equation}
where token embeddings are L2-normalized, making the dot product equivalent to cosine similarity.

The \textbf{backward contrastive loss} $\mathcal{L}_{\text{backward}}$ constrains the D$\to$Q retrieval path, compelling the model to align visual details with the query's intent. To maintain consistency with the MVA strategy in the forward pass, the backward loss is also computed across both the original and augmented image views. It is defined symmetrically using the generic loss function $\mathcal{L}(\cdot)$, where the image serves as the anchor:
\begin{equation}
    \label{eq:backward_loss_revised}
    \mathcal{L}_{\text{backward}} = \mathcal{L}(I_{\text{ori}}, Q_{\text{pos}}, \{Q_{\text{neg}}\}) + \beta \cdot \mathcal{L}(I^{\text{aug}}_{\text{ori}}, Q_{\text{pos}}, \{Q_{\text{neg}}\}),
\end{equation}
where $\{Q_{\text{neg}}\}$  is the set of hard negative queries generated via the HNQS pipeline. The same hyperparameter $\beta$ is used to balance the contribution of the augmented view.

The negative sets $\{I_{\text{neg}}\}$ and $\{Q_{\text{neg}}\}$ are central to Evo-Retriever. Crucially, the selection of specific negatives from these sets at each training step is governed by our LLM-EC strategy. Minimizing $\mathcal{L}_{total}$ drives Evo-Retriever to continually align visual and semantic representations, thereby enhancing generalization in multi-modal retrieval.

\subsection{{LLM-Guided Evolutionary Curriculum (LLM-EC)}.} 
\label{sec:llm_curriculum} 
\begin{figure}
	\centering
	\includegraphics[width=85mm]{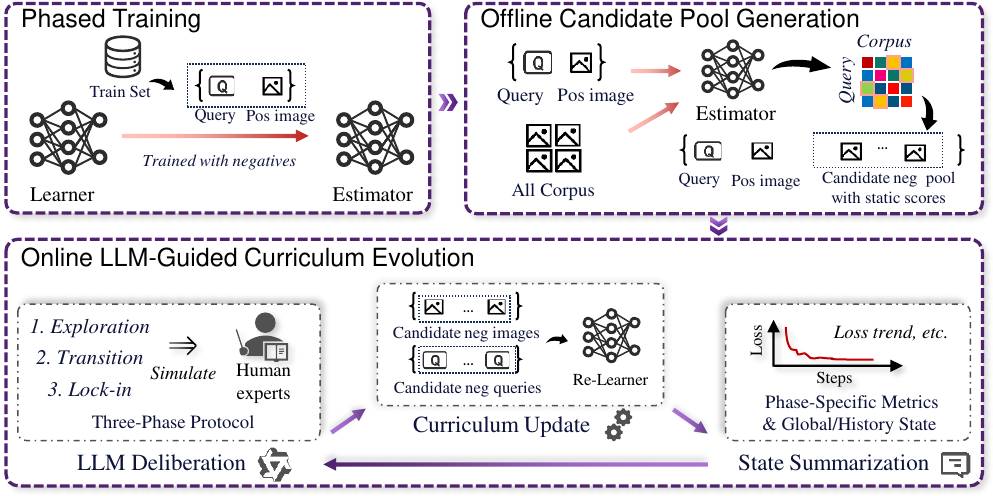}
	\caption{
		Overview of the proposed LLM-EC.
	}
	\label{fig_llm_overview}
\end{figure}

Existing hard-negative mining strategies, such as the Top-95 \% approach in NV-Retriever \cite{gabriel2024nv}, rely on static thresholds or pre-defined, \textit{state-agnostic} heuristics (e.g., linear interpolation). They fundamentally fail to adapt to the model’s evolving capabilities, leading to diminishing training signals and ultimately capping peak performance (\cref{fig_motivation}). To address this, we propose \textbf{LLM-EC}, a two-stage system comprising: (1)~\emph{Offline Candidate Pool Generation} and (2)~\emph{Online LLM-Guided Curriculum Evolution} (~\cref{fig_llm_overview}). This allows the model to alternate between estimator and learner, enabling continuous performance gains.

\paragraph{{Offline Candidate Pool Generation}.} 
\label{sec:Offline-Candidate-Pool-Generation} 
LLM-EC uses two types of hard negatives for the Viewpoint-Pathway Collaboration: hard-negative \emph{images} for the Q$\to$D path and hard negative \emph{queries} for the D$\to$Q path. To construct a high-quality set of hard negative images, we prefer a \textit{scheduled, pre-computed candidate pool} over on-the-fly discovery from limited samples during training. However, a candidate pool constructed by an external model may not accurately reflect the state of the model currently being trained. Therefore, we construct the candidate pool based on the current model’s own representation space.

First, we perform a brief warm-up phase using only in-batch negatives, allowing the model to establish an initial representation space for assessing sample difficulty. We then switch the model from learner to estimator and perform a one-time, offline global mining pass. Specifically, for each query $q$ in the training set, we retrieve and store the top-$N$ most similar negative documents from the entire corpus $\mathcal{D}$. This process yields a dedicated candidate pool $\mathcal{C}_q$ for each query, which forms the static \textbf{search space} for our dynamic curriculum. Notably, the hard negative queries for the D$\to$Q path are also pre-computed for curriculum scheduling. This design amortizes the high cost of global retrieval, enabling efficient online sample selection.

\paragraph{{Online LLM-Guided Curriculum Evolution}.} \label{sec:Online-LLM-Guided-Curriculum-Evolution} With a pre-computed candidate pool, a curriculum needs to assign negative samples of appropriate difficulty to the model. We formalize this by defining the curriculum as $M$ discrete difficulty intervals, each defined by a threshold range $[\tau_{\text{low}}, \tau_{\text{high}}]$ over a positive-aware difficulty measure (details in the Appendix). Following the strategy in NV-Retriever, this measure determines negative difficulty by its relative proximity to the corresponding positive. The task of selecting the appropriate interval (an `Action`) is delegated to an external LLM within a closed-loop feedback process. Unlike fixed-threshold schedulers, the controller conditions on structured summaries of recent training dynamics and history, enabling non-monotonic adjustments such as rolling back to easier intervals when instability emerges (see Appendix Sec.~\ref{sec:appendix_Trajectory}).

Our online curriculum evolution occurs during training and follows a three-step cycle: \textbf{(i) State Summarization:} Automatic aggregation of key metrics (e.g., mean hard-negative loss, loss trends) into a structured report. \textbf{(ii) LLM Deliberation:} The report is fed to the LLM meta-controller, which follows our Three-Phase Decision Protocol to select the next {`Action`}. The detailed protocol and example prompts are provided in the Appendix. \textbf{(iii) Curriculum Update:} The system adopts the new difficulty interval and mines hard negatives for the subsequent training phase. 

To ensure that LLM's decisions are both scientific and robust, a \textit{Three-Phase Decision Protocol} is proposed to simulate human curriculum design.

\textbf{Phase 1: Exploration.} Systematically map the "difficulty–performance landscape". The LLM explores a broad set of untested difficulty intervals while monitoring the loss, applying adaptive difficulty adjustments to escape regions of instability (excessively high loss) or stagnation (persistently low loss) and to promote efficient exploration.

\textbf{Phase 2: Transition.} Select an optimal "anchor point" for the main training stage. The LLM analyzes the full exploration history, filters for `Actions` that yielded "effective learning" (defined as loss within an ideal range, e.g., 0.3 <= loss <= 1.2), and chooses the most challenging among them to seed the next phase.

\textbf{Phase 3: Lock-in.} Conduct fine-grained adjustments during the main training phase. The LLM periodically assesses "learning velocity" by comparing mean loss at the start and end of each review period, and may (i)~\textbf{increase difficulty} by one step if mastery or substantial progress is detected, (ii)~\textbf{decrease difficulty} if the model struggles, or (iii) ~\textbf{maintain} the current level to ensure stable learning.

This structured protocol converts curriculum design from a heuristic task into a measurable, automated optimization. Under the new curriculum, the model’s role shifts from estimator to learner, thereby co-evolving with the curriculum’s difficulty.
\section{Experiments}
\label{sec:experiments}

%-----20251111-----
%% ====================================================================
%% 4.1 Datasets - No changes needed, format is standard.
%% ====================================================================
\subsection{Datasets}
Our training data consists of three components: (i) standard query-document pairs, (ii) synthetically generated hard negative queries, and (iii) dynamically mined hard negative document pages. 
The standard query-document pairs are a collection of approximately 480K instances from three datasets: ColPali~\cite{faysse2024colpali}, VisRAG-Ret-Train-Synthetic~\cite{yu2024visragvisionbasedretrievalaugmentedgeneration}, and VisRAG-Ret-Train-In-domain~\cite{yu2024visragvisionbasedretrievalaugmentedgeneration}. 
These datasets provide diverse and challenging multi-modal samples to enhance the model's generalization capability on visual retrieval tasks. 
Both the synthetic hard negative queries and the dynamically mined hard negatives are constructed based on these 480K standard pairs.
For validation, we use the official validation set of ColPali, which contains 500 samples.

\begin{table*}[t]
	\caption
	{	
		Performance comparison on the \textbf{ViDoRe V2} benchmark. All scores are nDCG@5 (\%). \textbf{Bold} and \underline{underlined} numbers denote the 1st and 2nd scores, respectively. Our models achieve \textbf{SOTA} performance, especially on challenging multilingual subsets.
	}
	\label{tab:vidore_results}
	\begin{center}
        % 调整字号
		\fontsize{9}{9}\selectfont
        % 调整行高（关键代码）：1.1 表示行高为默认的1.1倍，可按需增大
		\renewcommand{\arraystretch}{1.1}  
		%引入tabular环境
		\begin{tabular}{l|c|c|cccc}
			\specialrule{1.5pt}{0pt}{\belowrulesep}
			\multirow{2}{*}{Model} & \multirow{2}{*}{Size (B)}
			& \multirow{2}{*}{Avg.} & \multicolumn{1}{c}{\fontsize{8pt}{8pt}\selectfont MIT} & \multicolumn{1}{c}{\fontsize{8pt}{8pt}\selectfont Economics} & \multicolumn{1}{c}{\fontsize{8pt}{8pt}\selectfont ESG Restaurant} & \multicolumn{1}{c}{\fontsize{8pt}{8pt}\selectfont ESG Restaurant} \\
			
			% \cmidrule(r){4-5} \cmidrule(r){6-7} \cmidrule(r){8-10}
			
			& & & \fontsize{8pt}{8pt}\selectfont Biomedical & \fontsize{8pt}{8pt}\selectfont Macro & \fontsize{8pt}{8pt}\selectfont Human  &  
			\fontsize{8pt}{8pt}\selectfont Synthetic \\

			& & & \fontsize{8pt}{8pt}\selectfont Multilingual & \fontsize{8pt}{8pt}\selectfont Multilingual & \fontsize{8pt}{8pt}\selectfont English  &  
			\fontsize{8pt}{8pt}\selectfont  Multilingual \\
			
			%			\midrule
			\specialrule{0.5pt}{0.5pt}{1.5pt}
            \texttt{llama-nemoretriever-3b-v1} & 4.4 & \underline{63.5} & \underline{64.3} & 55.9 & 75.4 & \underline{58.6} \\
            \texttt{llama-nemoretriever-1b-v1} & 2.4 & 62.1 & 62.9 & 53.2 & \underline{76.4} & 55.9 \\
            \texttt{colnomic-embed-multimodal-7b} & 7.0 & 60.8 & 63.4 & 56.2 & 68.7 & 54.8 \\
            \texttt{colqwen2.5-v0.2} & 3.0 & 59.3 & 59.2 & 53.3 & 66.4 & 58.3 \\
            \texttt{jina-embeddings-v4} & 3.8 & 57.6 & 60.9 & 51.9 & 65.1 & 52.5 \\
            \texttt{colnomic-embed-multimodal-3b} & 3.0 & 55.5 & 62.5 & 53.3 & 57.0 & 49.2 \\
            \texttt{colqwen2-v1.0} & 2.2 & 55.0 & 56.3 & 50.6 & 60.4 & 52.5 \\
			\specialrule{0.5pt}{0.5pt}{1.5pt} %
            \textbf{Evo-Retriever-3B (Ours)} & 3.0 & {63.3} & {63.3} & \textbf{59.6} & {73.0} & {57.3} \\
            \textbf{Evo-Retriever-7B (Ours)} & 7.0 & \textbf{65.2} & \textbf{65.2} & \underline{59.1} & \textbf{77.0} & \textbf{59.7} \\
			\specialrule{1.5pt}{0.5pt}{0.5pt}
		\end{tabular}
	\end{center}
\end{table*}

\subsection{Implementation Details}

\textbf{Model Setup}: We adopt the Qwen-VL series~\cite{bai2025qwen2} as the backbone of our Evo-Retriever, utilizing two variants: Qwen2.5-VL-3B-Instruct and Qwen2.5-VL-7B-Instruct. Following the multi-vector late-interaction paradigm, the projection layer maps the final token embeddings to a 128-dimensional space. The maximum number of visual tokens per image is set to 1,024. Adhering to ColPali’s query augmentation strategy, we append five \texttt{<unused0>} special tokens to each query input sequence. {The $\alpha$ and $\beta$ in the loss function are set to 1 and 1, respectively.} The HNQS pipeline uses the Qwen2.5-VL-72B for negative query generation.

\textbf{Training Procedure}: Evo-Retriever achieves training optimization through a combination of phased training and LLM-EC. First, we train the model for one epoch using all 480k training pairs, utilizing only in-batch negative samples and the InfoNCE loss, and set the top-$N$ to 200 to build the candidate pool. Subsequently, the LLM meta-controller receives the model training state, generates a dynamic evolutionary curriculum, and retrains the model for one epoch. All models are fine-tuned with Low-Rank Adaptation (LoRA)~\cite{hu2022lora} at ranks 32 for both the 3B and 7B models. We use the \texttt{paged\_adamw\_8bit} optimizer with a learning rate of $2 \times 10^{-5}$, a cosine decay scheduler, and a warmup ratio of 2\%. The global batch size is set to 32. During each training step, every positive sample is paired with two synthesized hard negative queries and the two hard negative images with the highest similarity scores from the difficulty interval specified by the LLM-EC. All training is performed on 8 H20 GPUs using data parallelism.
%During each training step, every positive sample is paired with two hard negative documents and two synthesized hard negative queries.

% \textbf{LLM-guided curriculum}: The meta-controller uses the Qwen3-235B-A22B, accessible via its API. In the \textbf{curriculum phasing}, training begins with an \textit{Exploration} phase for the first 60 steps, during which the LLM-guided curriculum is updated every 2 steps. A \textit{Transition} phase occurs at step 61, where the model trains on the selected "anchor action" for 200 steps. Subsequently, the process enters the \textit{Lock-in} phase, with curriculum updates performed every 200 steps. 
\textbf{LLM-guided curriculum}: The meta-controller uses the Qwen3-235B-A22B, accessible via its API. The \textbf{curriculum phasing} involves a 60-step \textit{Exploration} (updates every 2 steps), a 200-step \textit{Transition} on a selected "anchor action", and a \textit{Lock-in} phase (curriculum updates every 200 steps). In the \textbf{decision protocol}, the LLM operates on a discrete action space of $M$=16 difficulty intervals, strictly following the \textit{Three-Phase Decision Protocol} detailed in Sec.~\ref{sec:llm_curriculum}. These are predefined overlapping similarity intervals ranging from [0.70, 0.85] to [0.95, 0.995]. The definitions for all 16 intervals are provided in the Appendix. LLM‑EC introduces only minor training overhead: the controller is queried periodically (about 100k tokens over the entire training process) and operates only during training, with no impact on inference efficiency. 
%The complete prompt is provided in the appendix. 
%Our custom trainer ensures that the LLM's decisions, made on the primary process, are broadcast and synchronized across all GPUs to maintain a consistent curriculum state.

\subsection{Evaluation}
To comprehensively evaluate performance, we conduct experiments on two representative public benchmarks: ViDoRe Benchmark V2~\cite{macé2025vidorebenchmarkv2raising}, a challenging multilingual evaluation set for assessing zero-shot cross-lingual capabilities, and the VisDoc task from MMEB~\cite{jiang2025vlm2vec}, which evaluates retrieval on structurally complex documents. Following official protocols, we report {nDCG@5} for ViDoRe V2 and {nDCG@linear 5} for MMEB (VisDoc).

\subsection{Main Results}

\paragraph{Results on ViDoRe V2.}
\Cref{tab:vidore_results} benchmarks performance on ViDoRe V2, with all scores reported as nDCG@5. While prior models like \texttt{colqwen2.5-v0.2} (59.3 \%) and \texttt{colnomic-embed-multimodal-7b} (60.8 \%) establish strong baselines, their performance indicates that advanced architectures alone are insufficient without equally advanced training strategies. The previous SOTA, \texttt{llama-nemoretriever-colembed-3b-v1} (63.5 \%) performs less favorably on challenging multilingual subsets (e.g., 55.9 \% on \textit{Economics Macro Multilingual}).

In contrast, our \textbf{Evo-Retriever-7B} sets a new SOTA with an overall score of \textbf{65.2 \%}, delivering a substantial +1.7 \% improvement.
Notably, our smaller \textbf{Evo-Retriever-3B} model also attains a high score of 63.3 \%, outperforming the architecturally-similar 3B model \texttt{colqwen2.5-v0.2} by 4.0 \%, demonstrating superior efficiency and generalization. 

\paragraph{Results on MMEB.}
\Cref{tab:mmeb_results_resized} shows the evaluation results on the MMEB (VisDoc) benchmark. This benchmark is complementary in design to ViDoRe V2: while ViDoRe V2 focuses on deep cross-lingual retrieval capabilities, MMEB assesses generalization \textit{breadth} and \textit{robustness} across diverse document structures and semantic domains by integrating multi-task datasets.

The results reveal that sophisticated training strategies are critical for top performance. For instance, \texttt{Seed-1.6-Embedding} (73.44 \%), despite its simpler single-vector architecture, surpasses the multi-vector \texttt{colpali-v1.3} (70.97 \%) by leveraging a multi-stage static curriculum. However, all these leading methods, including the strong \texttt{gme-Qwen2-VL-7B} (75.18 \%), are fundamentally constrained by their reliance on \textbf{pre-defined, static curricula} that are agnostic to the model's real-time learning state. 
In stark contrast, Evo-Retriever is powered by an \textbf{LLM meta-controller} that implements a state-aware, evolutionary curriculum. We attribute this gain primarily to the synergy between our representation enhancement modules and the adaptive LLM-guided curriculum. The significant performance gap over all static-curriculum baselines validates that our adaptive approach is a more effective and efficient paradigm for training SOTA retrievers.

\begin{table*}
	\caption
	{	
		Performance comparison on the \textbf{MMEB (VisDoc)} benchmark. All scores are nDCG@linear 5 (\%). \textbf{Bold} and \underline{underlined} numbers denote the 1st and 2nd scores, respectively. Our models demonstrate superior generalization and robustness across a wide range of document understanding tasks.
	}
	\label{tab:mmeb_results_resized}
	\begin{center}
        % 调整字号
		\fontsize{9}{9}\selectfont
        % 调整行高（关键代码）：1.1 表示行高为默认的1.1倍，可按需增大
		\renewcommand{\arraystretch}{1.1}  
		%引入tabular环境
		\begin{tabular}{l|c|c|cccc}
			\specialrule{1.5pt}{0pt}{\belowrulesep}
            
			Model & Size (B) & Avg. & ViDoRe V1 & ViDoRe V2 & VisRAG & VisDoc-OOD \\
			
			%			\midrule
			\specialrule{0.5pt}{0.5pt}{1.5pt}
            \texttt{gme-Qwen2-VL-7B-Instruct} & 8.3 & 75.18 & \textbf{89.44} & 55.61 & 84.99 & \textbf{44.40} \\
            \texttt{Seed-1.6-embedding} & - & 73.44 & 85.53 & 56.57 & 84.74 & 43.14 \\
            \texttt{gme-Qwen2-VL-2B-Instruct} & 2.2 & 72.71 & 86.15 & 53.96 & 82.52 & 43.12 \\
            \texttt{colpali-v1.3} & 2.9 & 70.97 & 83.60 & 51.98 & 81.13 & 43.12 \\
            \texttt{Ops-MM-embedding-v1-7B} & 8.3 & 70.34 & 80.05 & \underline{59.59} & 79.32 & 43.34 \\
            \texttt{Ops-MM-embedding-v1-2B} & 2.2 & 66.96 & 76.39 & 53.18 & 77.64 & 41.17 \\
            \texttt{VLM2Vec-V2.0-Qwen2VL-2B} & 2.2 & 65.36 & 75.52 & 44.86 & 79.38 & 39.43 \\
			\specialrule{0.5pt}{0.5pt}{1.5pt} %
            \textbf{Evo-Retriever-3B (Ours)}   & 3.0 & \underline{75.91} & {88.14} & {58.64} & \underline{89.06} & {42.84} \\
            \textbf{Evo-Retriever-7B (Ours)} & 7.0 & \textbf{77.12} & \underline{89.35} & \textbf{61.41} & \textbf{89.28} & \underline{44.00} \\
			\specialrule{1.5pt}{0.5pt}{0.5pt}
		\end{tabular}
	\end{center}
\end{table*}

\begin{table}
	\caption
	{	
		Ablation study on ViDoRe V2 (3B model). {\textbf{Bold} numbers denote the best scores}.
	}
	\label{tab:ablation_cvpr}
	\begin{center}
        % 调整字号
		\fontsize{9}{9}\selectfont
        % 调整行高（关键代码）：1.1 表示行高为默认的1.1倍，可按需增大
		\renewcommand{\arraystretch}{1.1}  
		%引入tabular环境
		\begin{tabular}{
				>{\centering}m{0.5cm}
				>{\centering}m{4cm}
				>{\centering\arraybackslash}m{1.5cm}}
			\specialrule{1.5pt}{0pt}{\belowrulesep}
            
			\textbf{ID} & \textbf{Configuration} & \textbf{nDCG@5} \\
			
			%			\midrule
			\specialrule{0.5pt}{0.5pt}{1.5pt}
            (a) & Baseline (\texttt{Net0}) & 61.17 \\
			\specialrule{0.5pt}{0.5pt}{1.5pt} %
            \multicolumn{3}{c}{\textit{--- Multi-view Alignment ---}} \\
            (b) & Downsample-only & 59.75 \\
            (c) & Stitched-only   & 58.59 \\
            (d) & + MVA (\texttt{Net1})& 62.25 \\
            \specialrule{0.5pt}{0.5pt}{1.5pt} %
            \multicolumn{3}{c}{\textit{--- Bidirectional Learning ---}} \\
            (e) & + BCL (\texttt{Net2})& 61.84 \\
            \specialrule{0.5pt}{0.5pt}{1.5pt} %
            \multicolumn{3}{c}{\textit{--- Combined Representation Backbone ---}} \\
            (f) & Net0 + MVA + BCL & 62.39
 \\
            \specialrule{0.5pt}{0.5pt}{1.5pt} %
            \multicolumn{3}{c}{\textit{--- Curriculum Strategies ---}} \\
            (g) & Fixed Top-K 90\%& 60.01\\
             (h) & Fixed Top-K 95\%&60.99\\
             (i) & Fixed Top-K 99.9\%&60.18\\
             (j) & Fixed Window 80-95\%&61.20\\
            (k) & Fixed Window 80-98\%& 62.10\\
            (l)& Rule-based Oracle & 62.81 \\
            (m)& LLM-EC (Ours)& 63.05 \\
            \specialrule{0.5pt}{0.5pt}{1.5pt} %
            \multicolumn{3}{c}{\textit{--- Overall Performance ---}} \\
            (n) & \textbf{Full Model (All components)} & \textbf{63.30}\\
			\specialrule{1.5pt}{0.5pt}{0.5pt}
		\end{tabular}
	\end{center}
\end{table}

\begin{table}[h!]
    \caption{Analysis of key design choices for LLM-EC, showing deviations from the default setting. {\textbf{Bold} numbers denote the best scores}.}
    \label{tab:hyperparam_ablation}
    \begin{center}
        % 调整字号
		\fontsize{9}{9}\selectfont
        % 调整行高
		\renewcommand{\arraystretch}{1.1}  
		% 引入tabular环境
		\begin{tabular}{
				>{\raggedright}m{1.5cm} % Factor (左对齐)
				>{\raggedright}m{4.0cm} % Setting (左对齐)
				>{\centering\arraybackslash}m{1.2cm}} % nDCG@5
			\specialrule{1.5pt}{0pt}{\belowrulesep}
            
			\textbf{Factor} & \textbf{Setting} & \textbf{nDCG@5} \\
			
			\specialrule{1.0pt}{0.5pt}{1.5pt}
			\multicolumn{3}{l}{\textit{1. Exploration Phase}} \\
			& Enabled (default, 60 steps) & 63.05 \\
			& {Disabled} & {61.86} \\
			\specialrule{0.5pt}{0.5pt}{1.5pt}

			\multicolumn{3}{l}{\textit{2. Difficulty Granularity}} \\
			& 10 Intervals (Coarse) & 62.89 \\
			& 16 Intervals (Default) & 63.05 \\
			& 21 Intervals (Fine) & 62.15 \\
			\specialrule{0.5pt}{0.5pt}{1.5pt}

			\multicolumn{3}{l}{\textit{3. LLM Meta-Controller Size}} \\
			&{Qwen3-235B-A22B (Default)} & 63.05 \\
			& {Qwen3-32B} & {\textbf{63.30}} \\
			\specialrule{1.5pt}{0.5pt}{0.5pt}
		\end{tabular}
	\end{center}
\end{table}

\subsection{Ablation Studies}
\label{sec:ablation}

% We systematically dissect the effectiveness of our Evo-Retriever framework through a comprehensive ablation study on the ViDoRe V2 benchmark using our 3B model, as detailed in \Cref{tab:ablation_cvpr}. Our baseline model (\textbf{Net0}) employs a standard InfoNCE loss with only in-batch negatives. All scores are reported as nDCG@5.
We conduct an ablation study on the ViDoRe V2 benchmark using our 3B model to dissect the contributions of each component. Our baseline (\textbf{Net0}) is a standard InfoNCE retriever with only in-batch negatives, achieving a score of 61.17\% (\cref{tab:ablation_cvpr}). All scores in this section are reported as nDCG@5.

\paragraph{Effectiveness of Representation Enhancement.}
Adding the MVA component alone boosts performance by +1.08 \% (61.17 \% → 62.25 \%). To validate its design, we test two ablations: \texttt{Downsample-only} (simplifying the augmented view) and \texttt{Stitched-only} (removing the original view).  Compared to the baseline, these simplified variants degrade performance to 59.75 \% (-1.42 \%) and 58.59 \% (-2.58 \%) respectively, confirming the necessity of our dual-view strategy. This design is crucial, as it preserves the high-fidelity global context via the original view while enforcing scale and orientation invariance through the augmented one.
%Their significant performance drops of -1.42\% and -2.58\% respectively confirm the necessity of our dual-view strategy. 
Next, we isolate \textbf{BCL} by applying a fixed hard-negative sampling strategy. Per positive pair, we randomly sample two hard negative queries (from HNQS pool), and sample two hard negative documents from a fixed 80-98 \% similarity window. This setup improves the baseline by +0.67 \%  (61.17 \% → 61.84 \%). When combined, these modules form our enhanced representation backbone (\textbf{Net0+MVA+BCL}), which improves the baseline by +1.22 \% (61.17 \% → 62.39 \%), underscoring the significant benefits of a multi-faceted representation enhancement strategy.

\paragraph{Impact of Curriculum Strategies.}
We next evaluate curriculum strategies on the  Q$\to$D path of the initial \textbf{Net0} baseline (61.17 \%).  
First, establishing a strong static curriculum—a Fixed Window (80-98\% similarity) for hard negative documents—elevates performance substantially to 62.10 \% (+0.93 \%) over the Net0 baseline. 
%We first establish the strongest static method: a \textbf{Fixed Window} (80-98 \%) at 62.10 \%, which significantly outperforms Fixed Top-K variants. 
This static baseline is surpassed by our dynamic, hard-coded \textbf{Rule-based Oracle} (62.81\%), a deterministic curriculum that adjusts difficulty using fixed loss thresholds. Finally, our LLM-EC pushes performance to a new high of 63.05\% (+0.24\%). Unlike the rule-based oracle that relies on fixed loss thresholds, LLM‑EC considers broader training dynamics (e.g., loss trends and history), enabling flexible adjustments such as proactive rollbacks when instability is detected. Qualitative trajectory comparisons are shown in Appendix Sec.~\ref{sec:appendix_Trajectory}.

%Finally, our \textbf{LLM-EC} reaches a superior 63.05\%. This performance gap highlights the LLM's ability to more robustly execute the intended expert logic, overcoming the inherent brittleness of a hard-coded system.

\paragraph{Analysis of LLM-EC Design Choices.}
\label{sec:ablation_analysis}
We now dissect LLM-EC's key design choices, with detailed results in \Cref{tab:hyperparam_ablation}. Our analysis reveals three critical factors for its success:

\textbf{The Exploration Phase is Crucial.} Disabling the initial Exploration phase, where the LLM probes the model's state to determine a suitable starting difficulty, causes a critical performance drop of -1.19\%  (63.05 \% → 61.86 \%). This suggests that adapting the starting difficulty to the model’s initial training state is more effective than using a pre-defined starting point.

\textbf{Difficulty Granularity Requires Balance.} The number of difficulty intervals is key. Performance degrades with too few intervals (e.g., 10 intervals, 62.89 \%) or too many (e.g., 21 intervals, 62.15 \%). The latter is likely due to insufficient convergence within each narrow band. Our choice of 16 intervals (63.05 \%) strikes an effective balance between curriculum precision and learning stability.

\textbf{Controller Scale: Robust Reasoning without Massive Scale.} Replacing the 235B controller with Qwen3‑32B matches the top score of 63.30 \% (\cref{tab:hyperparam_ablation}). This does not imply trivial logic: LLM‑EC still outperforms the rigid Rule‑based Oracle (\cref{tab:ablation_cvpr}) by flexibly interpreting the natural‑language protocol. These results suggest that massive controller scale is not necessary for effective protocol-guided curriculum control in our setting.

\section{Conclusion}

% -----20251112-----
% We propose Evo-Retriever, a model–curriculum co-evolutionary retrieval framework for complex visual documents. The Viewpoint–Pathway Collaboration leverages perspective expansion and confounding synthesis to enhance representations in both explicit spatial and implicit semantics. An external LLM acts as a meta-controller, scheduling difficult samples based on model state, maintaining information gradients while adapting course difficulty to model training. Evo-Retriever achieves SOTA results on ViDoRe V2 and MMEB VisDoc, with nDCG@5 of 65.2 \% and nDCG@linear-5 of 77.1 \%, demonstrating superior performance. Future work includes a lighter meta-controller, a reinforcement learning-based scheduler, etc., with the potential for broad integration into RAG and OCR-free document understanding pipelines.

We propose Evo‑Retriever, a model–curriculum co‑evolution framework for complex visual document retrieval. The Viewpoint–Pathway Collaboration improves spatial invariance and textual disambiguation, while an LLM meta‑controller adaptively schedules hard negatives according to training dynamics. Evo‑Retriever achieves SOTA performance on ViDoRe V2 and MMEB (VisDoc), reaching nDCG@5 scores of 65.2 \% and 77.1 \%.
{
    \small
    \bibliographystyle{ieeenat_fullname}
    \bibliography{main}

@String(CVPR= {IEEE Conf. Comput. Vis. Pattern Recog.})

@String(ICCV= {Int. Conf. Comput. Vis.})

@String(ICLR = {Int. Conf. Learn. Represent.})

@String(IJCAI = {IJCAI})

@String(CVPR  = {CVPR})

@String(ICCV  = {ICCV})

@String(ICLR  = {ICLR})

@inproceedings{khattab2020colbert,
  author       = {Omar Khattab and
                  Matei Zaharia},
  editor       = {Jimmy X. Huang and
                  Yi Chang and
                  Xueqi Cheng and
                  Jaap Kamps and
                  Vanessa Murdock and
                  Ji{-}Rong Wen and
                  Yiqun Liu},
  title        = {ColBERT: Efficient and Effective Passage Search via Contextualized
                  Late Interaction over {BERT}},
  booktitle    = {Proceedings of the 43rd International {ACM} {SIGIR} conference on
                  research and development in Information Retrieval, {SIGIR} 2020, Virtual
                  Event, China, July 25-30, 2020},
  pages        = {39--48},
  publisher    = {{ACM}},
  year         = {2020},
  url          = {https://doi.org/10.1145/3397271.3401075},
  doi          = {10.1145/3397271.3401075},
  bibsource    = {dblp computer science bibliography, https://dblp.org}
}

@inproceedings{radford2021learning,
  author       = {Alec Radford and
                  Jong Wook Kim and
                  Chris Hallacy and
                  Aditya Ramesh and
                  Gabriel Goh and
                  Sandhini Agarwal and
                  Girish Sastry and
                  Amanda Askell and
                  Pamela Mishkin and
                  Jack Clark and
                  Gretchen Krueger and
                  Ilya Sutskever},
  editor       = {Marina Meila and
                  Tong Zhang},
  title        = {Learning Transferable Visual Models From Natural Language Supervision},
  booktitle    = {Proceedings of the 38th International Conference on Machine Learning,
                  {ICML} 2021, 18-24 July 2021, Virtual Event},
  series       = {Proceedings of Machine Learning Research},
  volume       = {139},
  pages        = {8748--8763},
  publisher    = {{PMLR}},
  year         = {2021},
  url          = {http://proceedings.mlr.press/v139/radford21a.html},
  bibsource    = {dblp computer science bibliography, https://dblp.org}
}

@inproceedings{chen2022pali,
  author       = {Xi Chen and
                  Xiao Wang and
                  Soravit Changpinyo and
                  A. J. Piergiovanni and
                  Piotr Padlewski and
                  Daniel Salz and
                  Sebastian Goodman and
                  Adam Grycner and
                  Basil Mustafa and
                  Lucas Beyer and
                  Alexander Kolesnikov and
                  Joan Puigcerver and
                  Nan Ding and
                  Keran Rong and
                  Hassan Akbari and
                  Gaurav Mishra and
                  Linting Xue and
                  Ashish V. Thapliyal and
                  James Bradbury and
                  Weicheng Kuo},
  title        = {PaLI: {A} Jointly-Scaled Multilingual Language-Image Model},
  booktitle    = {The Eleventh International Conference on Learning Representations,
                  {ICLR} 2023, Kigali, Rwanda, May 1-5, 2023},
  publisher    = {OpenReview.net},
  year         = {2023},
  url          = {https://openreview.net/forum?id=mWVoBz4W0u},
  bibsource    = {dblp computer science bibliography, https://dblp.org}
}

@inproceedings{nassar2022tableformer,
  author       = {Ahmed S. Nassar and
                  Nikolaos Livathinos and
                  Maksym Lysak and
                  Peter W. J. Staar},
  title        = {TableFormer: Table Structure Understanding with Transformers},
  booktitle    = {{IEEE/CVF} Conference on Computer Vision and Pattern Recognition,
                  {CVPR} 2022, New Orleans, LA, USA, June 18-24, 2022},
  pages        = {4604--4613},
  publisher    = {{IEEE}},
  year         = {2022},
  url          = {https://doi.org/10.1109/CVPR52688.2022.00457},
  doi          = {10.1109/CVPR52688.2022.00457},
  bibsource    = {dblp computer science bibliography, https://dblp.org}
}

@inproceedings{pfitzmann2022doclaynet,
  author       = {Birgit Pfitzmann and
                  Christoph Auer and
                  Michele Dolfi and
                  Ahmed S. Nassar and
                  Peter W. J. Staar},
  editor       = {Aidong Zhang and
                  Huzefa Rangwala},
  title        = {DocLayNet: {A} Large Human-Annotated Dataset for Document-Layout Segmentation},
  booktitle    = {{KDD} '22: The 28th {ACM} {SIGKDD} Conference on Knowledge Discovery
                  and Data Mining, Washington, DC, USA, August 14 - 18, 2022},
  pages        = {3743--3751},
  publisher    = {{ACM}},
  year         = {2022},
  url          = {https://doi.org/10.1145/3534678.3539043},
  doi          = {10.1145/3534678.3539043},
  bibsource    = {dblp computer science bibliography, https://dblp.org}
}

@inproceedings{faysse2024colpali,
  author       = {Manuel Faysse and
                  Hugues Sibille and
                  Tony Wu and
                  Bilel Omrani and
                  Gautier Viaud and
                  C{\'{e}}line Hudelot and
                  Pierre Colombo},
  title        = {ColPali: Efficient Document Retrieval with Vision Language Models},
  booktitle    = {The Thirteenth International Conference on Learning Representations,
                  {ICLR} 2025, Singapore, April 24-28, 2025},
  publisher    = {OpenReview.net},
  year         = {2025},
  url          = {https://openreview.net/forum?id=ogjBpZ8uSi},
  bibsource    = {dblp computer science bibliography, https://dblp.org}
}

@inproceedings{ma2024unifying,
  author       = {Xueguang Ma and
                  Sheng{-}Chieh Lin and
                  Minghan Li and
                  Wenhu Chen and
                  Jimmy Lin},
  editor       = {Yaser Al{-}Onaizan and
                  Mohit Bansal and
                  Yun{-}Nung Chen},
  title        = {Unifying Multimodal Retrieval via Document Screenshot Embedding},
  booktitle    = {Proceedings of the 2024 Conference on Empirical Methods in Natural
                  Language Processing, {EMNLP} 2024, Miami, FL, USA, November 12-16,
                  2024},
  pages        = {6492--6505},
  publisher    = {Association for Computational Linguistics},
  year         = {2024},
  url          = {https://doi.org/10.18653/v1/2024.emnlp-main.373},
  doi          = {10.18653/V1/2024.EMNLP-MAIN.373},
  bibsource    = {dblp computer science bibliography, https://dblp.org}
}

@article{monsefi2024detailclip,
  title={Detailclip: Detail-oriented clip for fine-grained tasks},
  author={Monsefi, Amin Karimi and Sailaja, Kishore Prakash and Alilooee, Ali and Lim, Ser-Nam and Ramnath, Rajiv},
  journal={arXiv preprint arXiv:2409.06809},
  year={2024}
}

@article{liu2024textmonkey,
  title={Textmonkey: An ocr-free large multimodal model for understanding document},
  author={Liu, Yuliang and Yang, Biao and Liu, Qiang and Li, Zhang and Ma, Zhiyin and Zhang, Shuo and Bai, Xiang},
  journal={IEEE Transactions on Pattern Analysis and Machine Intelligence},
  year={2026},
  publisher={IEEE}
}

@article{xu2025llama,
  title={Llama Nemoretriever Colembed: Top-Performing Text-Image Retrieval Model},
  author={Xu, Mengyao and Moreira, Gabriel and Ak, Ronay and Osmulski, Radek and Babakhin, Yauhen and Yu, Zhiding and Schifferer, Benedikt and Oldridge, Even},
  journal={arXiv preprint arXiv:2507.05513},
  year={2025}
}

@inproceedings{nacson2025docvlm,
  author       = {Mor Shpigel Nacson and
                  Aviad Aberdam and
                  Roy Ganz and
                  Elad Ben{-}Avraham and
                  Alona Golts and
                  Yair Kittenplon and
                  Shai Mazor and
                  Ron Litman},
  title        = {DocVLM: Make Your {VLM} an Efficient Reader},
  booktitle    = {{IEEE/CVF} Conference on Computer Vision and Pattern Recognition,
                  {CVPR} 2025, Nashville, TN, USA, June 11-15, 2025},
  pages        = {29005--29015},
  publisher    = {Computer Vision Foundation / {IEEE}},
  year         = {2025},
  url          = {https://openaccess.thecvf.com/content/CVPR2025/html/Nacson\_DocVLM\_Make\_Your\_VLM\_an\_Efficient\_Reader\_CVPR\_2025\_paper.html},
  doi          = {10.1109/CVPR52734.2025.02701},
  bibsource    = {dblp computer science bibliography, https://dblp.org}
}

@article{macé2025vidorebenchmarkv2raising,
  title={Vidore benchmark v2: Raising the bar for visual retrieval},
  author={Mac{\'e}, Quentin and Loison, Ant{\'o}nio and Faysse, Manuel},
  journal={arXiv preprint arXiv:2505.17166},
  year={2025}
}

@inproceedings{karpukhin-etal-2020-dense,
    title = "Dense Passage Retrieval for Open-Domain Question Answering",
    author = "Karpukhin, Vladimir  and
      Oguz, Barlas  and
      Min, Sewon  and
      Lewis, Patrick  and
      Wu, Ledell  and
      Edunov, Sergey  and
      Chen, Danqi  and
      Yih, Wen-tau",
    editor = "Webber, Bonnie  and
      Cohn, Trevor  and
      He, Yulan  and
      Liu, Yang",
    booktitle = "Proceedings of the 2020 Conference on Empirical Methods in Natural Language Processing (EMNLP)",
    month = nov,
    year = "2020",
    address = "Online",
    publisher = "Association for Computational Linguistics",
    url = "https://aclanthology.org/2020.emnlp-main.550/",
    doi = "10.18653/v1/2020.emnlp-main.550",
    pages = "6769--6781"
}

@inproceedings{caffagni2025recurrence,
  author       = {Davide Caffagni and
                  Sara Sarto and
                  Marcella Cornia and
                  Lorenzo Baraldi and
                  Rita Cucchiara},
  title        = {Recurrence-Enhanced Vision-and-Language Transformers for Robust Multimodal
                  Document Retrieval},
  booktitle    = {{IEEE/CVF} Conference on Computer Vision and Pattern Recognition,
                  {CVPR} 2025, Nashville, TN, USA, June 11-15, 2025},
  pages        = {9286--9295},
  publisher    = {Computer Vision Foundation / {IEEE}},
  year         = {2025},
  url          = {https://openaccess.thecvf.com/content/CVPR2025/html/Caffagni\_Recurrence-Enhanced\_Vision-and-Language\_Transformers\_for\_Robust\_Multimodal\_Document\_Retrieval\_CVPR\_2025\_paper.html},
  doi          = {10.1109/CVPR52734.2025.00867},
  bibsource    = {dblp computer science bibliography, https://dblp.org}
}

@inproceedings{gunther2025jina,
    title = "jina-embeddings-v4: Universal Embeddings for Multimodal Multilingual Retrieval",
    author = {G{\"u}nther, Michael  and
      Sturua, Saba  and
      Akram, Mohammad Kalim  and
      Mohr, Isabelle  and
      Ungureanu, Andrei  and
      Wang, Bo  and
      Eslami, Sedigheh  and
      Martens, Scott  and
      Werk, Maximilian  and
      Wang, Nan  and
      Xiao, Han},
    editor = "Adelani, David Ifeoluwa  and
      Arnett, Catherine  and
      Ataman, Duygu  and
      Chang, Tyler A.  and
      Gonen, Hila  and
      Raja, Rahul  and
      Schmidt, Fabian  and
      Stap, David  and
      Wang, Jiayi",
    booktitle = "Proceedings of the 5th Workshop on Multilingual Representation Learning (MRL 2025)",
    month = nov,
    year = "2025",
    address = "Suzhuo, China",
    publisher = "Association for Computational Linguistics",
    url = "https://aclanthology.org/2025.mrl-main.36/",
    doi = "10.18653/v1/2025.mrl-main.36",
    pages = "531--550",
    ISBN = "979-8-89176-345-6"
}

@article{zhang2024gme,
  title={GME: Improving Universal Multimodal Retrieval by Multimodal LLMs},
  author={Zhang, Xin and Zhang, Yanzhao and Xie, Wen and Li, Mingxin and Dai, Ziqi and Long, Dingkun and Xie, Pengjun and Zhang, Meishan and Li, Wenjie and Zhang, Min},
  journal={arXiv preprint arXiv:2412.16855},
  year={2024}
}

@inproceedings{jiang2025vlm2vec,
  author       = {Ziyan Jiang and
                  Rui Meng and
                  Xinyi Yang and
                  Semih Yavuz and
                  Yingbo Zhou and
                  Wenhu Chen},
  title        = {VLM2Vec: Training Vision-Language Models for Massive Multimodal Embedding
                  Tasks},
  booktitle    = {The Thirteenth International Conference on Learning Representations,
                  {ICLR} 2025, Singapore, April 24-28, 2025},
  publisher    = {OpenReview.net},
  year         = {2025},
  url          = {https://openreview.net/forum?id=TE0KOzWYAF},
  bibsource    = {dblp computer science bibliography, https://dblp.org}
}

@article{bai2025qwen2,
  title={Qwen2.5-VL Technical Report},
  author={Bai, Shuai and Chen, Keqin and Liu, Xuejing and Wang, Jialin and Ge, Wenbin and Song, Sibo and Dang, Kai and Wang, Peng and Wang, Shijie and Tang, Jun and others},
  journal={arXiv preprint arXiv:2502.13923},
  year={2025}
}

@inproceedings{gabriel2024nv,
  author       = {Gabriel de Souza Pereira Moreira and
                  Radek Osmulski and
                  Mengyao Xu and
                  Ronay Ak and
                  Benedikt Schifferer and
                  Even Oldridge},
  editor       = {Meeyoung Cha and
                  Chanyoung Park and
                  Noseong Park and
                  Carl Yang and
                  Senjuti Basu Roy and
                  Jessie Li and
                  Jaap Kamps and
                  Kijung Shin and
                  Bryan Hooi and
                  Lifang He},
  title        = {Improving Text Embedding Models with Positive-aware Hard-negative
                  Mining},
  booktitle    = {Proceedings of the 34th {ACM} International Conference on Information
                  and Knowledge Management, {CIKM} 2025, Seoul, Republic of Korea, November
                  10-14, 2025},
  pages        = {2169--2178},
  publisher    = {{ACM}},
  year         = {2025},
  url          = {https://doi.org/10.1145/3746252.3761254},
  doi          = {10.1145/3746252.3761254},
  bibsource    = {dblp computer science bibliography, https://dblp.org}
}

@inproceedings{yu2024visragvisionbasedretrievalaugmentedgeneration,
  author       = {Shi Yu and
                  Chaoyue Tang and
                  Bokai Xu and
                  Junbo Cui and
                  Junhao Ran and
                  Yukun Yan and
                  Zhenghao Liu and
                  Shuo Wang and
                  Xu Han and
                  Zhiyuan Liu and
                  Maosong Sun},
  title        = {VisRAG: Vision-based Retrieval-augmented Generation on Multi-modality
                  Documents},
  booktitle    = {The Thirteenth International Conference on Learning Representations,
                  {ICLR} 2025, Singapore, April 24-28, 2025},
  publisher    = {OpenReview.net},
  year         = {2025},
  url          = {https://openreview.net/forum?id=zG459X3Xge},
  bibsource    = {dblp computer science bibliography, https://dblp.org}
}

@inproceedings{wasserman2025docrerank,
  author       = {Navve Wasserman and
                  Oliver Heinimann and
                  Yuval Golbari and
                  Tal Zimbalist and
                  Eli Schwartz and
                  Michal Irani},
  editor       = {Christos Christodoulopoulos and
                  Tanmoy Chakraborty and
                  Carolyn Rose and
                  Violet Peng},
  title        = {DocReRank: Single-Page Hard Negative Query Generation for Training
                  Multi-Modal {RAG} Rerankers},
  booktitle    = {Proceedings of the 2025 Conference on Empirical Methods in Natural
                  Language Processing, {EMNLP} 2025, Suzhou, China, November 4-9, 2025},
  pages        = {8640--8658},
  publisher    = {Association for Computational Linguistics},
  year         = {2025},
  url          = {https://doi.org/10.18653/v1/2025.emnlp-main.436},
  doi          = {10.18653/V1/2025.EMNLP-MAIN.436},
  bibsource    = {dblp computer science bibliography, https://dblp.org}
}

@inproceedings{hu2022lora,
  author       = {Edward J. Hu and
                  Yelong Shen and
                  Phillip Wallis and
                  Zeyuan Allen{-}Zhu and
                  Yuanzhi Li and
                  Shean Wang and
                  Lu Wang and
                  Weizhu Chen},
  title        = {LoRA: Low-Rank Adaptation of Large Language Models},
  booktitle    = {The Tenth International Conference on Learning Representations, {ICLR}
                  2022, Virtual Event, April 25-29, 2022},
  publisher    = {OpenReview.net},
  year         = {2022},
  url          = {https://openreview.net/forum?id=nZeVKeeFYf9},
  bibsource    = {dblp computer science bibliography, https://dblp.org}
}

@article{xu2025multi,
  title={A Multi-Granularity Retrieval Framework for Visually-Rich Documents},
  author={Xu, Mingjun and Wang, Zehui and Cai, Hengxing and Zhong, Renxin},
  journal={arXiv preprint arXiv:2505.01457},
  year={2025}
}

@article{lee2024unified,
  title={Unified Multimodal Interleaved Document Representation for Retrieval},
  author={Lee, Jaewoo and Ko, Joonho and Baek, Jinheon and Jeong, Soyeong and Hwang, Sung Ju},
  journal={arXiv preprint arXiv:2410.02729},
  year={2024}
}

@article{meng2025vlm2vec,
  title={Vlm2vec-v2: Advancing multimodal embedding for videos, images, and visual documents},
  author={Meng, Rui and Jiang, Ziyan and Liu, Ye and Su, Mingyi and Yang, Xinyi and Fu, Yuepeng and Qin, Can and Chen, Zeyuan and Xu, Ran and Xiong, Caiming and others},
  journal={arXiv preprint arXiv:2507.04590},
  year={2025}
}

@inproceedings{wang2025vidorag,
  author       = {Qiuchen Wang and
                  Ruixue Ding and
                  Zehui Chen and
                  Weiqi Wu and
                  Shihang Wang and
                  Pengjun Xie and
                  Feng Zhao},
  editor       = {Christos Christodoulopoulos and
                  Tanmoy Chakraborty and
                  Carolyn Rose and
                  Violet Peng},
  title        = {ViDoRAG: Visual Document Retrieval-Augmented Generation via Dynamic
                  Iterative Reasoning Agents},
  booktitle    = {Proceedings of the 2025 Conference on Empirical Methods in Natural
                  Language Processing, {EMNLP} 2025, Suzhou, China, November 4-9, 2025},
  pages        = {9113--9134},
  publisher    = {Association for Computational Linguistics},
  year         = {2025},
  url          = {https://doi.org/10.18653/v1/2025.emnlp-main.464},
  doi          = {10.18653/V1/2025.EMNLP-MAIN.464},
  bibsource    = {dblp computer science bibliography, https://dblp.org}
}

@article{qin2025unimoco,
  title={UniMoCo: Unified Modality Completion for Robust Multi-Modal Embeddings},
  author={Qin, Jiajun and Pu, Yuan and He, Zhuolun and Kim, Seunggeun and Pan, David Z and Yu, Bei},
  journal={arXiv preprint arXiv:2505.11815},
  year={2025}
}

@article{zhao2024controllable,
  title={Controllable dense captioner with multimodal embedding bridging},
  author={Zhao, Yuzhong and Liu, Yue and Guo, Zonghao and Wu, Weijia and Gong, Chen and Ye, Qixiang and Wan, Fang},
  journal={arXiv preprint arXiv:2401.17910},
  year={2024}
}

@article{zhang2020deep,
  author       = {Yifan Zhang and
                  Wengang Zhou and
                  Min Wang and
                  Qi Tian and
                  Houqiang Li},
  title        = {Deep Relation Embedding for Cross-Modal Retrieval},
  journal      = {{IEEE} Trans. Image Process.},
  volume       = {30},
  pages        = {617--627},
  year         = {2021},
  url          = {https://doi.org/10.1109/TIP.2020.3038354},
  doi          = {10.1109/TIP.2020.3038354},
  bibsource    = {dblp computer science bibliography, https://dblp.org}
}

@article{sourati2025lad,
  title={LAD-RAG: Layout-aware Dynamic RAG for Visually-Rich Document Understanding},
  author={Sourati, Zhivar and Wang, Zheng and Liu, Marianne Menglin and Hu, Yazhe and Guo, Mengqing and Bharadwaj, Sujeeth and Han, Kyu and Sheng, Tao and Ravi, Sujith and Dehghani, Morteza and others},
  journal={arXiv preprint arXiv:2510.07233},
  year={2025}
}

@article{mahadevkar2024exploring,
  author       = {Supriya V. Mahadevkar and
                  Shruti Patil and
                  Ketan Kotecha and
                  Lim Way Soong and
                  Tanupriya Choudhury},
  title        = {Exploring AI-driven approaches for unstructured document analysis
                  and future horizons},
  journal      = {J. Big Data},
  volume       = {11},
  number       = {1},
  pages        = {92},
  year         = {2024},
  url          = {https://doi.org/10.1186/s40537-024-00948-z},
  doi          = {10.1186/S40537-024-00948-Z},
  bibsource    = {dblp computer science bibliography, https://dblp.org}
}

@article{wang2025vrag,
  title={VRAG-RL: Empower Vision-Perception-Based RAG for Visually Rich Information Understanding via Iterative Reasoning with Reinforcement Learning},
  author={Wang, Qiuchen and Ding, Ruixue and Zeng, Yu and Chen, Zehui and Chen, Lin and Wang, Shihang and Xie, Pengjun and Huang, Fei and Zhao, Feng},
  journal={arXiv preprint arXiv:2505.22019},
  year={2025}
}

@inproceedings{biswas2021graph,
  author       = {Sanket Biswas and
                  Pau Riba and
                  Josep Llad{\'{o}}s and
                  Umapada Pal},
  editor       = {Elisa H. Barney Smith and
                  Umapada Pal},
  title        = {Graph-Based Deep Generative Modelling for Document Layout Generation},
  booktitle    = {Document Analysis and Recognition, {ICDAR} 2021 Workshops, Lausanne,
                  Switzerland, September 5-10, 2021, Proceedings, Part {II}},
  series       = {Lecture Notes in Computer Science},
  volume       = {12917},
  pages        = {525--537},
  publisher    = {Springer},
  year         = {2021},
  url          = {https://doi.org/10.1007/978-3-030-86159-9\_38},
  doi          = {10.1007/978-3-030-86159-9\_38},
  bibsource    = {dblp computer science bibliography, https://dblp.org}
}

@inproceedings{huang2025visual,
  author       = {Xin Huang and
                  Ruibin Li and
                  Tong Jia and
                  Wei Zheng and
                  Ya Wang},
  title        = {Visual Perturbation and Adaptive Hard Negative Contrastive Learning
                  for Compositional Reasoning in Vision-Language Models},
  booktitle    = {Proceedings of the Thirty-Fourth International Joint Conference on
                  Artificial Intelligence, {IJCAI} 2025, Montreal, Canada, August 16-22,
                  2025},
  pages        = {5435--5443},
  publisher    = {ijcai.org},
  year         = {2025},
  url          = {https://doi.org/10.24963/ijcai.2025/605},
  doi          = {10.24963/IJCAI.2025/605},
  bibsource    = {dblp computer science bibliography, https://dblp.org}
}

@inproceedings{fu2023learning,
  author       = {Zheren Fu and
                  Zhendong Mao and
                  Yan Song and
                  Yongdong Zhang},
  title        = {Learning Semantic Relationship among Instances for Image-Text Matching},
  booktitle    = {{IEEE/CVF} Conference on Computer Vision and Pattern Recognition,
                  {CVPR} 2023, Vancouver, BC, Canada, June 17-24, 2023},
  pages        = {15159--15168},
  publisher    = {{IEEE}},
  year         = {2023},
  url          = {https://doi.org/10.1109/CVPR52729.2023.01455},
  doi          = {10.1109/CVPR52729.2023.01455},
  bibsource    = {dblp computer science bibliography, https://dblp.org}
}

@article{teiletche2025modernvbert,
  title={ModernVBERT: Towards Smaller Visual Document Retrievers},
  author={Teiletche, Paul and Mac{\'e}, Quentin and Conti, Max and Loison, Antonio and Viaud, Gautier and Colombo, Pierre and Faysse, Manuel},
  journal={arXiv preprint arXiv:2510.01149},
  year={2025}
}

@inproceedings{he2023capstone,
  author       = {Xingwei He and
                  Yeyun Gong and
                  A{-}Long Jin and
                  Hang Zhang and
                  Anlei Dong and
                  Jian Jiao and
                  Siu{-}Ming Yiu and
                  Nan Duan},
  editor       = {Houda Bouamor and
                  Juan Pino and
                  Kalika Bali},
  title        = {{CAPSTONE:} Curriculum Sampling for Dense Retrieval with Document
                  Expansion},
  booktitle    = {Proceedings of the 2023 Conference on Empirical Methods in Natural
                  Language Processing, {EMNLP} 2023, Singapore, December 6-10, 2023},
  pages        = {10531--10541},
  publisher    = {Association for Computational Linguistics},
  year         = {2023},
  url          = {https://doi.org/10.18653/v1/2023.emnlp-main.651},
  doi          = {10.18653/V1/2023.EMNLP-MAIN.651},
  bibsource    = {dblp computer science bibliography, https://dblp.org}
}

@inproceedings{kwak2025qure,
  author       = {Jaehyun Kwak and
                  Ramahdani Muhammad Izaaz Inhar and
                  Se{-}Young Yun and
                  Sung{-}Ju Lee},
  editor       = {Aarti Singh and
                  Maryam Fazel and
                  Daniel Hsu and
                  Simon Lacoste{-}Julien and
                  Felix Berkenkamp and
                  Tegan Maharaj and
                  Kiri Wagstaff and
                  Jerry Zhu},
  title        = {QuRe: Query-Relevant Retrieval through Hard Negative Sampling in Composed
                  Image Retrieval},
  booktitle    = {Forty-second International Conference on Machine Learning, {ICML}
                  2025, Vancouver, BC, Canada, July 13-19, 2025},
  series       = {Proceedings of Machine Learning Research},
  volume       = {267},
  publisher    = {{PMLR} / OpenReview.net},
  year         = {2025},
  url          = {https://proceedings.mlr.press/v267/kwak25a.html},
  bibsource    = {dblp computer science bibliography, https://dblp.org}
}

@inproceedings{jang2023difficulty,
  author       = {Taeuk Jang and
                  Xiaoqian Wang},
  title        = {Difficulty-Based Sampling for Debiased Contrastive Representation
                  Learning},
  booktitle    = {{IEEE/CVF} Conference on Computer Vision and Pattern Recognition,
                  {CVPR} 2023, Vancouver, BC, Canada, June 17-24, 2023},
  pages        = {24039--24048},
  publisher    = {{IEEE}},
  year         = {2023},
  url          = {https://doi.org/10.1109/CVPR52729.2023.02302},
  doi          = {10.1109/CVPR52729.2023.02302},
  bibsource    = {dblp computer science bibliography, https://dblp.org}
}

@article{oord2018representation,
  title={Representation learning with contrastive predictive coding},
  author={Oord, Aaron van den and Li, Yazhe and Vinyals, Oriol},
  journal={arXiv preprint arXiv:1807.03748},
  year={2018}
}

@article{lan2025ume,
  title={UME-R1: Exploring Reasoning-Driven Generative Multimodal Embeddings},
  author={Lan, Zhibin and Niu, Liqiang and Meng, Fandong and Zhou, Jie and Su, Jinsong},
  journal={arXiv preprint arXiv:2511.00405},
  year={2025}
}

@inproceedings{yu2025cafe,
  author       = {Hao Yu and
                  Zhuokai Zhao and
                  Shen Yan and
                  Lukasz Korycki and
                  Jianyu Wang and
                  Baosheng He and
                  Jiayi Liu and
                  Lizhu Zhang and
                  Xiangjun Fan and
                  Hanchao Yu},
  title        = {{CAFE:} Unifying Representation and Generation with Contrastive-Autoregressive
                  Finetuning},
  booktitle    = {{IEEE/CVF} International Conference on Computer Vision, {ICCV} 2025
                  - Workshops, Honolulu, HI, USA, October 19-20, 2025},
  pages        = {6345--6356},
  publisher    = {{IEEE}},
  year         = {2025},
  url          = {https://doi.org/10.1109/ICCVW69036.2025.00659},
  doi          = {10.1109/ICCVW69036.2025.00659},
  bibsource    = {dblp computer science bibliography, https://dblp.org}
}

@article{gu2025unime,
  title={Unime-v2: Mllm-as-a-judge for universal multimodal embedding learning},
  author={Gu, Tiancheng and Yang, Kaicheng and Zhang, Kaichen and An, Xiang and Feng, Ziyong and Zhang, Yueyi and Cai, Weidong and Deng, Jiankang and Bing, Lidong},
  journal={arXiv preprint arXiv:2510.13515},
  year={2025}
}

@article{xue2025improve,
  title={Improve Multi-Modal Embedding Learning via Explicit Hard Negative Gradient Amplifying},
  author={Xue, Youze and Li, Dian and Liu, Gang},
  journal={arXiv preprint arXiv:2506.02020},
  year={2025}
}

@inproceedings{suresh2024cornstack,
  author       = {Tarun Suresh and
                  Revanth Gangi Reddy and
                  Yifei Xu and
                  Zach Nussbaum and
                  Andriy Mulyar and
                  Brandon Duderstadt and
                  Heng Ji},
  title        = {CoRNStack: High-Quality Contrastive Data for Better Code Retrieval
                  and Reranking},
  booktitle    = {The Thirteenth International Conference on Learning Representations,
                  {ICLR} 2025, Singapore, April 24-28, 2025},
  publisher    = {OpenReview.net},
  year         = {2025},
  url          = {https://openreview.net/forum?id=iyJOUELYir},
  bibsource    = {dblp computer science bibliography, https://dblp.org}
}
}

\clearpage
\setcounter{page}{1}
\maketitlesupplementary

\section{Negative Query Generation Prompt}
\label{sec:NegativeQueryGenerationPrompt}
% \begin{promptbox}
% \ttfamily % 等宽字体
% ''' \\
% You are given the following question:
% "question" \\
% \\
% The document can answer this question. \\
% \\
% Now, write 6 new questions that are:\\
% - Related to the topic,\\
% - Seem reasonable,\\
% - But cannot be answered using the document.\\
% \\
% These questions should require knowledge that is not in the document. \\
% \\
% Do not rephrase the original.\\
% \\
% Give exactly 6 new questions. Just list them:\\
% Variant 1: ...\\
% Variant 2: ...\\
% Variant 3: ...\\
% Variant 4: ...\\
% Variant 5: ...\\
% Variant 6: ...\\
% '''\\
% \end{promptbox}

\begin{promptbox}[Negative Query Generation Prompts]
\label{prompt:real-world-delib}
    You are given the following question:\\
    \{question\} \\
    \\
    The image can answer this question. \\
    \\
    Now, write 20 new questions that are:\\
    - Related to the topic,\\
    - Seem reasonable,\\
    - But cannot be answered using the image.\\
    \\
    These questions should require knowledge that is not in the image. \\
    \\
    Do not rephrase the original.\\
    \\
    Give exactly 20 new questions. Just list them:\\
    Variant 1: ...\\
    Variant 2: ...\\
    Variant 3: ...\\
    Variant 4: ...\\
    Variant 5: ...\\
    Variant 6: ...\\
    ...\\
    % '''
\end{promptbox}   % <--- 这里是修改过的地方！

% % 灰色背景框（通过基础命令实现，不依赖tcolorbox等宏包）
% \noindent% 取消首行缩进
% \colorbox{gray!15}{% 灰色背景（15%灰度）
%   \minipage{\dimexpr\linewidth-2\fboxsep\relax}% 自适应宽度（减去内边距）
%     \ttfamily% 等宽字体
%     % '''\\
%     You are given the following question:\\
%     \{question\} \\
%     \\
%     The image can answer this question. \\
%     \\
%     Now, write 20 new questions that are:\\
%     - Related to the topic,\\
%     - Seem reasonable,\\
%     - But cannot be answered using the image.\\
%     \\
%     These questions should require knowledge that is not in the image. \\
%     \\
%     Do not rephrase the original.\\
%     \\
%     Give exactly 20 new questions. Just list them:\\
%     Variant 1: ...\\
%     Variant 2: ...\\
%     Variant 3: ...\\
%     Variant 4: ...\\
%     Variant 5: ...\\
%     Variant 6: ...\\
%     ...\\
%     % '''
%   \endminipage%
% }

\section{Rationale and Design of  LLM-EC}
\label{sec:appendix_rationale}

As stated in \cref{sec:llm_curriculum} 
, the LLM-EC framework relies on a set of $M$ discrete difficulty intervals, defined over a positive-aware difficulty measure. The design of these intervals is critical. A naive uniform partitioning would ignore the highly non-uniform learning signals inherent in contrastive learning. We therefore adopt a \textbf{principled, non-uniform partitioning strategy}. This section details the theoretical and empirical rationale for this design, beginning with a gradient-sensitivity analysis that reveals the underlying problem structure.

\subsection{Gradient-Sensitivity Analysis of the Contrastive Loss}
\label{sec:appendix_gradient_analysis}

The informativeness of a negative sample is closely related to the gradient it induces during training. To formalize this, we analyze the gradient of our softplus-based margin loss (\cref{eq:margin_loss_definition}). As the total loss is a sum over individual negative samples, we can analyze the gradient contribution from a single negative, $I_{\text{neg}}^{(j)}$. The gradient of the loss with respect to its similarity score, $s_{\text{neg}}^{(j)}$, is:
\begin{equation}
\label{eq:gradient_formula_revised}
    \frac{\partial \mathcal{L}}{\partial s_{\text{neg}}^{(j)}} = \frac{1}{\tau} \sigma\left(\frac{\Delta s^{(j)}}{\tau}\right)
\end{equation}
where $\Delta s^{(j)} = s_{\text{neg}}^{(j)} - s_{\text{pos}}$ is the similarity gap, $\tau$ is the temperature, and $\sigma(\cdot)$ is the Sigmoid function. This equation reveals that the gradient magnitude is proportional to a sigmoid function of the similarity gap, scaled by $1/\tau$.

\begin{figure}[h!] % Use [h!] for "here if possible"
	\centering
	\includegraphics[width=0.7\columnwidth]{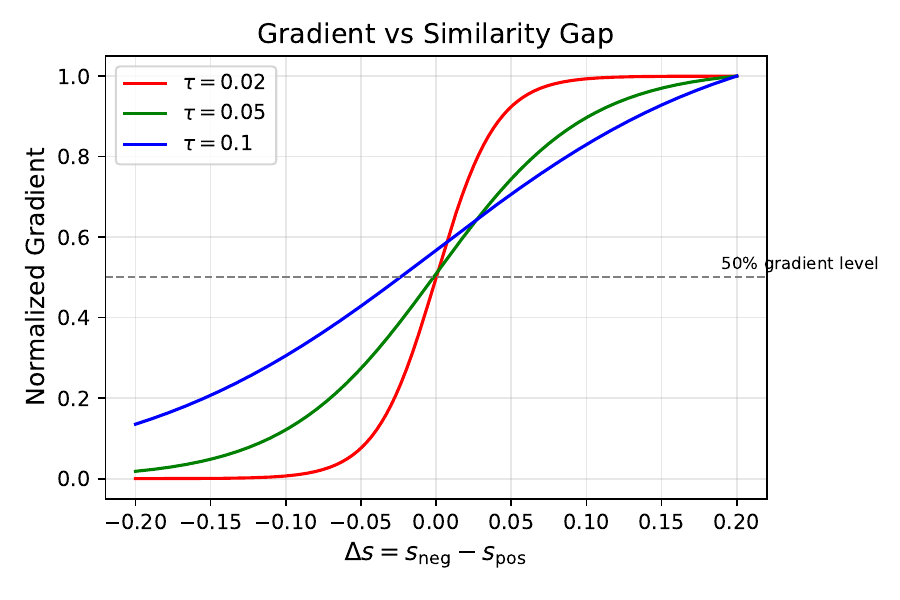} % Use relative width for better scaling
	\caption{
		\textbf{Impact of Temperature $\tau$ on the Normalized Gradient Profile.} The plot shows the normalized gradient, $\sigma(\Delta s / \tau)$, as a function of the similarity gap $\Delta s$. A lower $\tau$ (e.g., our setting of 0.02, red line) creates a much steeper and narrower transition zone where gradients become substantial. This highlights the need for a curriculum that can precisely target negatives within this narrow \enquote{sweet spot} to ensure efficient learning.
	}
	\label{fig:gradient_analysis} % Use fig: prefix for figures
\end{figure}

The relationship between the gradient and the similarity gap is visualized in \cref{fig:gradient_analysis}. The temperature $\tau$ critically governs the sharpness of this gradient profile. Our use of a low temperature ($\tau=0.02$, red line) results in a highly localized gradient response. The gradient is negligible for large negative $\Delta s$ (i.e., for easy negatives) but rises steeply as $\Delta s$ approaches zero. This indicates that the most informative negatives are concentrated in a narrow window around the decision boundary, where the loss provides strong yet still meaningful training signals. A curriculum must therefore be able to precisely identify and sample from this \enquote{sweet spot} to avoid wasting computation on uninformative samples and ensure efficient training. This principle motivates our design of a non-uniform, high-resolution action space for the LLM-EC.

\subsection{Rationale for Non-Uniform Partitioning}
\label{sec:nonuniform_partitioning}
Although the analysis in \cref{sec:appendix_gradient_analysis} is expressed in terms of the similarity gap $\Delta s$, it provides the theoretical motivation for our practical interval design based on a positive-aware difficulty measure, which captures the relative proximity of a negative to its paired positive. The design of the LLM-EC action space involves two core steps: defining the operating range and partitioning it. 

\textbf{Operating Range.} Our ablation study in \cref{tab:ablation_cvpr} directly informs the choice of range. Introducing a lower bound (e.g., \enquote{Fixed Window 80--95 \%}, ID j) improves performance to \textbf{61.20 \%} from \textbf{60.99 \%} (+0.21 \%) compared to including easy negatives (\enquote{Fixed Top-K 95\%}, ID h). Conversely, an excessively high upper bound (\enquote{Top-K 99.9 \%}, ID i) is detrimental, causing performance to drop to \textbf{60.18 \%} (-0.81 \%). To ensure a diverse action space for dynamic adjustment while avoiding these extremes, and guided by the gradient profile (\cref{sec:appendix_gradient_analysis}), we establish the primary operating range of the positive-aware difficulty measure as $[0.70, 0.995]$.

Within this range, the gradient response (\cref{fig:gradient_analysis}) is highly non-uniform. The gradient $\sigma(\Delta s/\tau)$ is both \textbf{negligible and insensitive} to the similarity gap for $\Delta s < -0.10$, but rises sharply as $s_{\text{neg}} \!\to\! s_{\text{pos}}$. A uniform partitioning would waste resolution on this low-signal zone while undersampling the narrow, high-information \enquote{sweet spot}.

Therefore, we define three zones based on the gradient curve: the \textbf{Low-Signal Zone} ($[0.70,0.85]$), the \textbf{Effective-Learning Zone} ($[0.85,0.98]$), and the \textbf{High-Risk Zone} ($[0.98,0.995]$). We allocate the $M=16$ intervals with a higher density in the Effective-Learning Zone (8 intervals) and a coarser density in the others (4 each). The final boundaries are determined via empirical quantiles of the data distribution (\cref{tab:difficulty_intervals_v1}), ensuring both reproducibility and data-informed alignment.

\begin{table}[t]
\centering
\caption{The action space of $M=16$ overlapping sampling policies.}
\label{tab:difficulty_intervals_v1}
\begin{tabular}{cl}
\toprule
\textbf{Action ID} & \textbf{Difficulty Range} \\
\midrule
\multicolumn{2}{l}{\textit{Low-Signal Zone }} \\
A & $[0.70, 0.85]$ \\
B & $[0.70, 0.90]$ \\
C & $[0.70, 0.92]$ \\
D & $[0.75, 0.90]$ \\
\midrule
\multicolumn{2}{l}{\textit{Effective-Learning Zone}} \\
E & $[0.75, 0.92]$ \\
F & $[0.75, 0.94]$ \\
G & $[0.80, 0.92]$ \\
H & $[0.80, 0.94]$ \\
I & $[0.80, 0.95]$ \\
J  & $[0.85, 0.96]$ \\
K & $[0.85, 0.97]$ \\
L & $[0.85, 0.98]$ \\
\midrule
\multicolumn{2}{l}{\textit{High-Risk Zone }} \\
M & $[0.90, 0.985]$ \\
N & $[0.92, 0.985]$ \\
O & $[0.95, 0.99]$ \\
P & $[0.95, 0.995]$ \\
\bottomrule
\end{tabular}
\end{table}

% This principled, heuristic partitioning forms the foundation for our curriculum's difficulty intervals.
\subsection{Formal Decision Protocol}
\label{app:formal_protocol}

The curriculum evolution in LLM-EC is governed by a deterministic, phase-dependent decision protocol. This protocol maps the training state to specific, reproducible difficulty-adjustment actions. The numerical parameters within the protocol are not arbitrary but are based on the following design principles and were parameterized via preliminary experiments.

\paragraph{Protocol Design and Parameterization.}
\begin{itemize}
    \item \textbf{Effective Learning Window ($[0.3, 1.2]$):} This is the empirically determined ideal range for the loss. For the softplus-based contrastive loss with a temperature of $\tau=0.02$, a loss below 0.3 heuristically indicates that negatives are too easy, providing insufficient learning signal. A loss above 1.2 suggests a risk of training instability, where the model struggles to converge. This window filters for actions that are both challenging and stable.

\item \textbf{Trend Estimation Window (20\%):}
To estimate learning velocity, the protocol compares the loss at the beginning ($\mathcal{L}_{\text{start}}$) and end ($\mathcal{L}_{\text{end}}$) of each review period. Both metrics are computed as averages over the first and last 20\% of steps within that period, respectively. This window provides a principled balance---wide enough to suppress batch-level variance, yet narrow enough to remain sensitive to short-term trends.

\item \textbf{Anchor Point Principle (Difficulty-Optimized Selection):} 
During the \textbf{Transition phase}, the controller applies a hierarchical selection strategy to identify the anchor that will seed the subsequent \textbf{Lock-in phase}. The goal is to choose the most challenging curriculum that still maintains stable learning dynamics, as observed during Exploration. If no such action exists, this indicates a potential curriculum calibration failure, rather than a case where an unsupported anchor should be forced. In this case, the system should re-examine the difficulty intervals, loss thresholds, or candidate pool, and may optionally perform an additional round of exploration before entering the Lock-in phase.  This two-level mechanism forms the core of the Phase~2 procedure.

\end{itemize}

\paragraph{Key Metric Definitions.}
\begin{itemize}
    \item \textbf{Action ($a$):} An integer index from 0 to 15, corresponding to a predefined difficulty interval (see \cref{tab:difficulty_intervals_v1}). In logs, these may be represented by letters \enquote{A} through \enquote{P}.
    \item \textbf{Hard Negative Loss ($\mathcal{L}_{\text{neg}}$):} The primary real-time performance metric for decision-making. It is the mean value of the \textbf{total contrastive loss, $\mathcal{L}_{total}$ (Eq.~\ref{eq:overall_loss})}, computed within a review window. This loss directly reflects the model's performance on the current curriculum's hard negatives. For historical records, this value is stored as \texttt{avg\_loss}.
    \item \textbf{Historical Summary ($H$):} A record of all previously executed actions. Each entry contains at least the action taken and its resulting average loss, i.e., a tuple of (\texttt{action}, \texttt{avg\_loss}), along with metadata such as the global step. This summary serves as the historical context for the overall decision-making process.
\end{itemize}

\paragraph{Phase 1: Exploration.}
\textbf{Objective:} To systematically test different difficulty intervals and map the difficulty-performance landscape.
\textbf{Decision Logic:}
\begin{enumerate}
    \item \textbf{High-Loss Anomaly:} If $\mathcal{L}_{\text{neg}} > 1.2$, difficulty is reduced by 2 intervals ($a_{\text{next}} \leftarrow \max(a_{\text{current}} - 2, 0)$).
    \item \textbf{Low-Loss Anomaly:} If $\mathcal{L}_{\text{neg}} < 0.05$ for \textbf{two consecutive} reviews, difficulty is increased by 3 intervals ($a_{\text{next}} \leftarrow \min(a_{\text{current}} + 3, 15)$).
    \item \textbf{Default Progression:} Otherwise, select the lowest-indexed action greater than the current one that has not been used in the last three reviews.
\end{enumerate}

\paragraph{Phase 2: Transition.}
\textbf{Objective:} To select the most robust \enquote{anchor} action for the next phase based on the exploration history.
\textbf{Decision Logic:}
\begin{enumerate}
    \item \textbf{Filter for Effective Actions:} Construct a set of valid actions, $A_{\text{valid}}$, by filtering history $H$ for actions whose \texttt{a.avg\_loss} falls within the Effective Learning Window of $[0.3, 1.2]$.
    \item \textbf{Select Optimal Anchor:} From the actions in $A_{\text{valid}}$, identify the one with the highest difficulty (i.e., the one with the largest Action ID, from \enquote{A} to \enquote{P}). This action is selected as the anchor to seed the Lock-in phase.
    \item \textbf{Fallback:} If $A_{\text{valid}}$ is empty, this is treated as a curriculum calibration failure rather than forcing an unsupported anchor choice. The system should re-examine the difficulty intervals, loss thresholds, or candidate pool, and optionally perform an additional exploration round before entering the Lock-in phase.
\end{enumerate}

\paragraph{Phase 3: Lock-in.}
\textbf{Objective:} To conduct fine-grained difficulty adjustments around the anchor.
\textbf{Decision Logic:} The protocol assesses \enquote{learning velocity} by comparing $\mathcal{L}_{\text{start}}$ and $\mathcal{L}_{\text{end}}$, calculated over the Trend Estimation Window.
\begin{enumerate}
    \item \textbf{Upgrade Condition:} Increase difficulty by 1 step if either is met:
        \begin{itemize}
            \item \textit{Absolute Mastery:} $\mathcal{L}_{\text{end}} < 0.3$.
            \item \textit{Significant Progress:} A relative loss reduction of $\ge 50\%$.
        \end{itemize}
    \item \textbf{Downgrade Condition:} Decrease difficulty by 1 step if there is a relative loss increase of $\ge 30\%$.
    \item \textbf{Maintain Condition:} If neither of the above is met, maintain the current action.
\end{enumerate}

\subsection{Real-World Deliberation Example}

This section presents a real-world example of the model’s end-to-end deliberation. In this case, the controller is in the \textbf{EXPLORATION} phase and must select the next difficulty interval based on the current interval, recent loss, and the last three actions. The trace below shows how it applies the exploration protocol and correctly selects action \textbf{C}.

\begin{promptbox}[Real-World Deliberation]
\label{prompt:real-world-delib} 
<thinking> \\
Current phase: EXPLORATION \\

Objective: Systematically explore the difficulty space by selecting a higher difficulty action that has not been used in the last three reviews, unless an anomaly rule is triggered. \\

Step 1: Identify the current action and recent history.\\
- Current interval: [0.7, 0.90]\\
- Current action: B\\
- Last 3 actions: \\
\quad Step 34: B\\
\quad Step 32: D\\
\quad Step 30: F\\

Therefore, the recent action set is \{B, D, F\}, and these actions should be avoided.\\

Step 2: Check anomaly conditions.\\
- hard\_negative\_loss\_mean = 0.3983\\
- High-loss anomaly condition: $L_{neg} > 1.2$ \\
\quad Not triggered, since $0.3983 < 1.2$.\\
- Low-loss anomaly condition: $L_{neg} < 0.05$ for two consecutive reviews \\
\quad Not triggered.\\

Therefore, follow the default exploration progression rule.\\

Step 3: Apply default progression.\\
The protocol specifies selecting the lowest-indexed action greater than the current one that has not been used in the last three reviews.\\
- Current action = B\\
- Higher actions begin from C\\
- C is not in \{B, D, F\}\\

Thus, C is the first valid higher action.\\

Step 4: Final decision.\\
Select action C as the next exploration step.\\
</thinking>\\

<answer>C</answer>.\\
\end{promptbox}

\section{Additional Ablation Studies}
\label{sec:appendix_ablation}

\subsection{Effect of the Temperature $\tau$}
\label{sec:appendix_tau}

We study the effect of the temperature $\tau$ in the margin loss (Eq.\eqref{eq:margin_loss_definition}) on ViDoRe V2 using Qwen2.5-VL-3B-Instruct under the same training setup as in our LLM-EC experiments  (\cref{sec:ablation_analysis}; all hyperparameters and data unchanged, only varying $\tau$). As shown in \cref{tab:tau_ablation}, a large temperature ($\tau = 1.0$) noticeably hurts performance, yielding 57.37 \% nDCG@5, whereas our default $\tau = 0.02$ achieves a better result (57.37 \% $\rightarrow$ 63.05 \%, +5.68 \%), indicating that a sharper margin is important for separating hard negatives in the multi-vector space.

\begin{table}[h]
    \caption{
    Ablation on the temperature $\tau$ in the margin loss on ViDoRe V2 (3B model). All scores are nDCG@5 (\%). \textbf{Bold} numbers denote the best score.
    }
    \label{tab:tau_ablation}
    \begin{center}
        \fontsize{9}{9}\selectfont
        \renewcommand{\arraystretch}{1.1}
        \begin{tabular}{c|c}
            \specialrule{1.5pt}{0pt}{\belowrulesep}
            $\tau$ & nDCG@5 \\
            \specialrule{0.5pt}{0.5pt}{1.5pt}
            1.0   & 57.37 \\
            \textbf{0.02 (Ours)} & \textbf{63.05} \\
            \specialrule{1.5pt}{0.5pt}{0.5pt}
        \end{tabular}
    \end{center}
\end{table}
\subsection{Effect of Interval Density: Non-uniform vs. Uniform Partitioning}
\label{sec:appendix_interval_density}

The design rationale for our \textbf{non-uniform, overlapping} difficulty partitioning is detailed in \cref{sec:nonuniform_partitioning}. Briefly, intervals are denser in the high-information Effective-Learning Zone ($[0.85,0.98]$) and coarser in the Low-Signal and High-Risk zones, with partial overlap to ensure smooth difficulty transitions.

Here, we isolate the effect of the density strategy by comparing our default design against a \textbf{uniform, overlapping} baseline. This baseline divides the full operating range $[0.700,\,0.995]$ into 16 equal-width intervals, each with fixed overlap between adjacent bins; all other LLM-EC settings remain identical. 

We evaluate both strategies on ViDoRe V2 using the Qwen2.5-VL-3B-Instruct model under the same setup as in our LLM-EC ablation experiments (\cref{sec:ablation_analysis}; all hyperparameters and data unchanged, only varying the partitioning strategy). Results are shown in \cref{tab:interval_density}.

\begin{table}[h]
    \caption{
    Effect of interval density on ViDoRe V2 (3B model). All scores are nDCG@5 (\%). \textbf{Bold} marks the best result.
    }
    \label{tab:interval_density}
    \begin{center}
        \fontsize{9}{9}\selectfont
        \renewcommand{\arraystretch}{1.1}
        \begin{tabular}{l|c}
            \specialrule{1.5pt}{0pt}{\belowrulesep}
            Partitioning Strategy & nDCG@5 \\
            \specialrule{0.5pt}{0.5pt}{1.5pt} 
            Uniform + Overlap & 61.89 \\
            \textbf{Non-uniform + Overlap (Ours)} & \textbf{63.05} \\
            \specialrule{1.5pt}{0.5pt}{0.5pt}
        \end{tabular}
    \end{center}
\end{table}

Our non-uniform approach improves performance by +1.16 \% absolute over the uniform-overlap baseline. This gain supports our design choice in \cref{sec:nonuniform_partitioning}: allocating higher resolution where hard negatives are concentrated enables the LLM meta-controller to make finer, more effective curriculum adjustments, while avoiding wasted granularity in low-signal regions.

\subsection{Transition and Lock-in Phase Lengths}
\label{sec:appendix_phase_length}

We vary the Transition and Lock-in phase lengths in the Three-Phase Decision Protocol (\cref{sec:llm_curriculum}) while fixing the Exploration phase to 60 steps. As shown in \cref{tab:phase_length}, sweeping Transition/Lock-in lengths from 50 to 200 steps shows that short phases (50 / 50) slightly underperform, whereas longer phases (200 / 200) achieve the best nDCG@5 (62.18 \% → 63.05 \%, +0.87 \%), indicating that moderately longer phases provide a more stable and informative curriculum signal.

\begin{table}[h]
    \caption{
    Ablation on Transition and Lock-in phase lengths for LLM-EC (3B, ViDoRe V2). All scores are nDCG@5 (\%). \textbf{Bold} numbers denote the best score.
    }
    \label{tab:phase_length}
    \begin{center}
        \fontsize{9}{9}\selectfont
        \renewcommand{\arraystretch}{1.1}
        \begin{tabular}{c|c|c}
            \specialrule{1.5pt}{0pt}{\belowrulesep}
            Transition steps & Lock-in steps & nDCG@5 \\
            \specialrule{0.5pt}{0.5pt}{1.5pt}
            50  & 50   & 62.18\\
            \textbf{200 (Ours)} & \textbf{200 (Ours)} & \textbf{63.05} \\
            \specialrule{1.5pt}{0.5pt}{0.5pt}
        \end{tabular}
    \end{center}
\end{table}

\subsection{Robustness and Transferability of LLM‑EC}
\label{sec:appendix_qwen4b}
We further analyze the robustness and generalization capability of LLM-EC under different settings.

\textbf{Backbone Transfer.}
To validate whether the curriculum strategy generalizes across 
backbones, we apply LLM‑EC to \textbf{Qwen3-VL-4B-Instruct}. Consistent with Sec.~\ref{sec:ablation_analysis}, each curriculum is evaluated on the \textbf{Q$\rightarrow$D path}, with only the hard negative pool being rebuilt. We additionally include a \textbf{Linear Curriculum} baseline that linearly schedules the action ID from the easiest (Action~`A') to the most challenging (Action~`P') over training. As shown in \cref{tab:qwen4b_llmec}, LLM-EC achieves the best performance, improving the baseline by +1.17 \% (64.03 \% $\rightarrow$ 65.20 \%). It also surpasses the strongest hard-coded baseline, \textbf{Rule-based Oracle} (64.36 \%), by +0.84 \%. These results indicate that, under the same curriculum protocol, LLM-EC can successfully adapt the scheduling decisions to a new backbone without additional tuning, whereas rule-based schedulers fail to achieve comparable improvements.

\begin{table}[t]
    \caption{
    Curriculum ablation on Qwen3-VL-4B-Instruct. All scores are nDCG@5 (\%). \textbf{Bold} numbers denote the best score.
    }
    \label{tab:qwen4b_llmec}
    \begin{center}
        \fontsize{9}{9}\selectfont
        \renewcommand{\arraystretch}{1.1}
        \begin{tabular}{l|c}
            \specialrule{1.5pt}{0pt}{\belowrulesep}
            Configuration & nDCG@5 \\
            \specialrule{0.5pt}{0.5pt}{1.5pt}
            baseline (Net0) & 64.03 \\
            Fixed Window 80--98\% & 64.32 \\
            Linear Curriculum & 64.29 \\
            Rule-based Oracle & 64.36 \\
            \textbf{LLM-EC (Ours)} & \textbf{65.20} \\ 
            \specialrule{1.5pt}{0.5pt}{0.5pt}
        \end{tabular}
    \end{center}
\end{table}

\textbf{Robustness to Threshold Perturbations.}
A potential concern is that LLM-EC may merely execute the threshold heuristics defined in the protocol. To test this, we perturb the thresholds used in the Upgrade and Downgrade Conditions while keeping all other training settings unchanged. 
Specifically, we relax the Absolute Mastery threshold from $L_{\mathrm{end}} < 0.3$ to $L_{\mathrm{end}} < 0.36$, increase the required relative loss reduction in the Upgrade Condition from $50\%$ to $60\%$, and decrease the relative loss increase threshold in the Downgrade Condition from $30\%$ to $25.5\%$. As shown in \cref{tab:threshold_perturb}, the Rule-based Oracle suffers a drop of $-1.06\%$ (62.81\% $\rightarrow$ 61.75\%), whereas LLM-EC maintains more stable performance with only a $-0.65\%$ drop (63.05\% $\rightarrow$ 62.40\%). The rule-based scheduler is sensitive to threshold shifts because its decisions are entirely threshold-driven, whereas LLM-EC adapts curriculum decisions in response to training dynamics. This indicates that the controller does not rely solely on fixed heuristic triggers.
\begin{table}[t]
    \caption{
    Robustness to threshold perturbations on ViDoRe V2 (Evo-Retriever-3B).
    In the Lock-in phase, thresholds in the Upgrade and Downgrade Conditions are perturbed: $L_{\mathrm{end}} < 0.3 \rightarrow 0.36$, relative loss reduction for upgrade $50\% \rightarrow 60\%$, and relative loss increase for downgrade $30\% \rightarrow 25.5\%$. Drop = Default $-$ Perturbed. All scores are nDCG@5 (\%).
    }
    \label{tab:threshold_perturb}
    \begin{center}
        \fontsize{9}{9}\selectfont
        \renewcommand{\arraystretch}{1.1}
        \begin{tabular}{l|c|c|c}
            \specialrule{1.5pt}{0pt}{\belowrulesep}
            Scheduler & Default & Perturbed & Drop \\
            \specialrule{0.5pt}{0.5pt}{1.5pt}
            Rule-based Oracle   & 62.81 & 61.75 & $-1.06$ \\
            \textbf{LLM-EC (Ours)} & \textbf{63.05} & \textbf{62.40} & $\mathbf{-0.65}$ \\
            \specialrule{1.5pt}{0.5pt}{0.5pt}
        \end{tabular}
    \end{center}
\end{table}

\subsection{Curriculum Trajectory Analysis}
\label{sec:appendix_Trajectory}
To better understand the behavior of LLM-EC, we visualize the curriculum difficulty trajectories during training under the same setting as the LLM-EC ablation experiments in \cref{sec:ablation_analysis} (Qwen2.5-VL-3B-Instruct backbone with a Qwen3-32B controller). Compared with the rule-based oracle, which follows monotonic threshold-triggered difficulty jumps, LLM-EC adapts its decisions based on training dynamics and may roll back to easier intervals when instability is detected. This trend-aware behavior helps maintain training within an effective learning regime.

\begin{figure}[ht]
\centering
\includegraphics[width=1\linewidth]{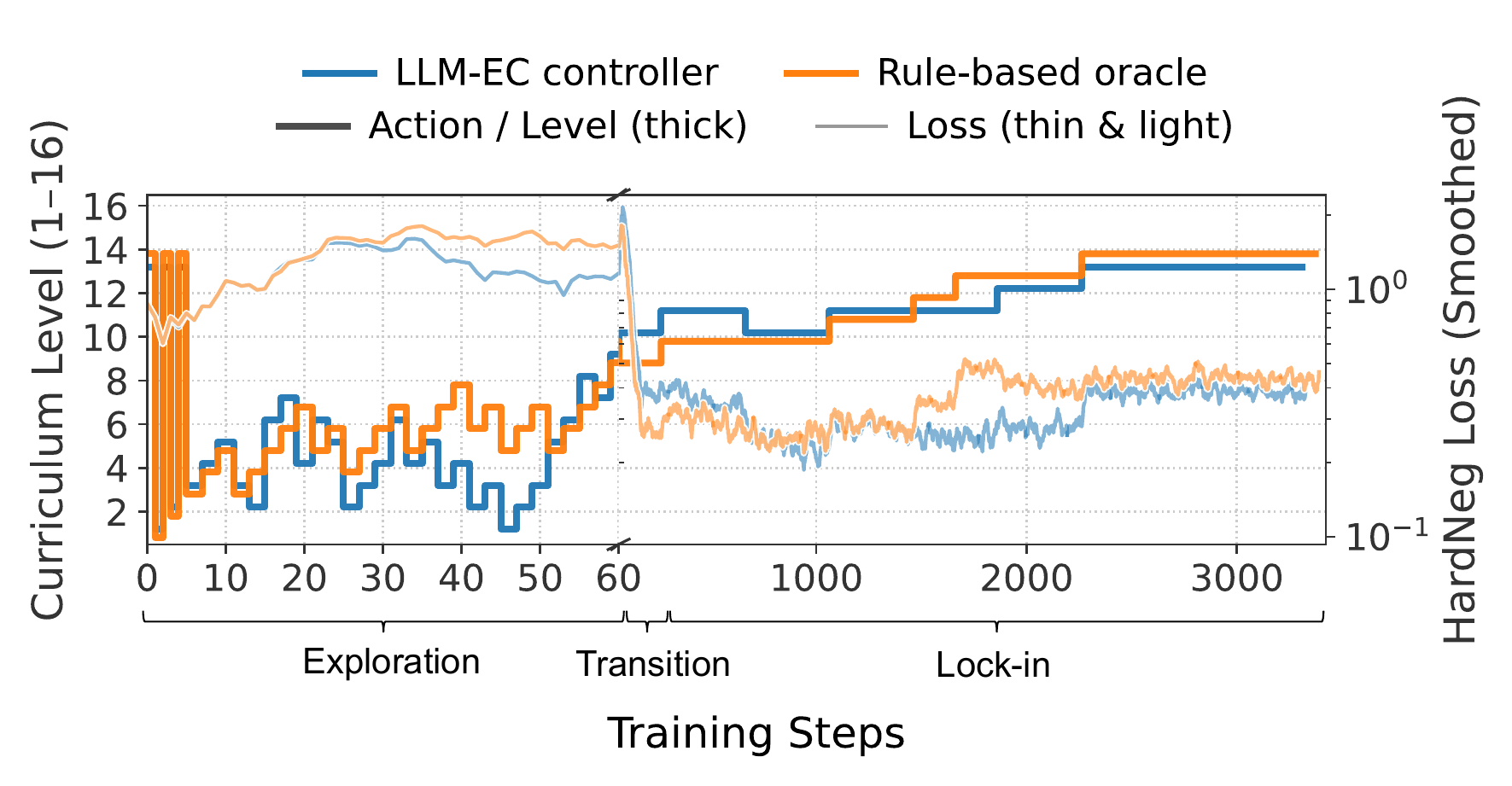}
\caption{Curriculum trajectories during training for Evo-Retriever-3B. Thick lines indicate the selected difficulty interval, while thin lines show the hard-negative loss. The rule-based oracle follows monotonic threshold-triggered jumps, whereas LLM-EC adapts difficulty according to training dynamics and performs rollback when instability is detected.}
\label{fig:onecol}
\end{figure}

\end{document}